\documentclass[10pt,twocolumn,letterpaper]{article}

\usepackage{cvpr}              % To produce the CAMERA-READY version
\usepackage{graphicx}
\usepackage{amsmath}
\usepackage{amssymb}
\usepackage{booktabs}
\usepackage{stfloats}
\usepackage{multirow}

\usepackage[pagebackref,breaklinks,colorlinks]{hyperref}

\usepackage[accsupp]{axessibility}  % Improves PDF readability for those with disabilities.

\usepackage[capitalize]{cleveref}
\crefname{section}{Sec.}{Secs.}
\Crefname{section}{Section}{Sections}
\Crefname{table}{Table}{Tables}
\crefname{table}{Tab.}{Tabs.}

\newcommand{\bigzero}{\mbox{\normalfont\bfseries 0}}

\def\cvprPaperID{0072} % *** Enter the CVPR Paper ID here
\def\confName{CVPR}
\def\confYear{2023}

\begin{document}

\title{NIKI: Neural Inverse Kinematics with Invertible Neural Networks\\for 3D Human Pose and Shape Estimation}

\author{{Jiefeng Li}$^{1*}$ \quad {Siyuan Bian}$^{1*}$ \quad Qi Liu$^1$ \quad Jiasheng Tang$^3$ \quad Fan Wang$^3$ \quad Cewu Lu$^{12}$\footnotemark[2] \\
$^1$Department of Computer Science and Engineering, Shanghai Jiao Tong University\\
$^2$MoE Key Lab of Artificial Intelligence, AI Institute, Shanghai Jiao Tong University\\
$^3$Alibaba Group \\
}
\maketitle

%%%%%%%%% ABSTRACT
\begin{abstract}
   With the progress of 3D human pose and shape estimation, state-of-the-art methods can either be robust to occlusions or obtain pixel-aligned accuracy in non-occlusion cases. However, they cannot obtain robustness and mesh-image alignment at the same time. In this work, we present NIKI (\textbf{N}eural \textbf{I}nverse \textbf{K}inematics with \textbf{I}nvertible Neural Network), which models bi-directional errors to improve the robustness to occlusions and obtain pixel-aligned accuracy. NIKI can learn from both the forward and inverse processes with invertible networks. In the inverse process, the model separates the error from the plausible 3D pose manifold for a robust 3D human pose estimation. In the forward process, we enforce the zero-error boundary conditions to improve the sensitivity to reliable joint positions for better mesh-image alignment. Furthermore, NIKI emulates the analytical inverse kinematics algorithms with the twist-and-swing decomposition for better interpretability. Experiments on standard and occlusion-specific benchmarks demonstrate the effectiveness of NIKI, where we exhibit robust and well-aligned results simultaneously. Code is available at \href{https://github.com/Jeff-sjtu/NIKI}{https://github.com/Jeff-sjtu/NIKI}.
\end{abstract}

\footnotetext[1]{Equal contribution.}
\footnotetext[2]{Cewu Lu is the corresponding author. He is the member of Qing Yuan Research Institute, Qi Zhi Institute and MoE Key Lab of Artificial Intelligence, AI Institute, Shanghai Jiao Tong University, China.}

%%%%%%%%% BODY TEXT
\section{Introduction}
\label{sec:intro}

Recovering 3D human pose and shape (HPS) from monocular input is a challenging problem. It has many applications\cite{peng2021neural,xiu2022icon,xiu2022econ,yi2022mime,dai2023sloper4d,liao2023car,yi2022generating}. Despite the rapid progress powered by deep neural networks~\cite{hmr,spin,vibe,zhang2021pymaf,li2021hybrik,kocabas2021pare}, the performance of existing methods is not satisfactory in complex real-world applications where people are often occluded and truncated by themselves, each other, and objects.

\begin{figure}[t]
   \centering
   \includegraphics[width=0.96\linewidth]{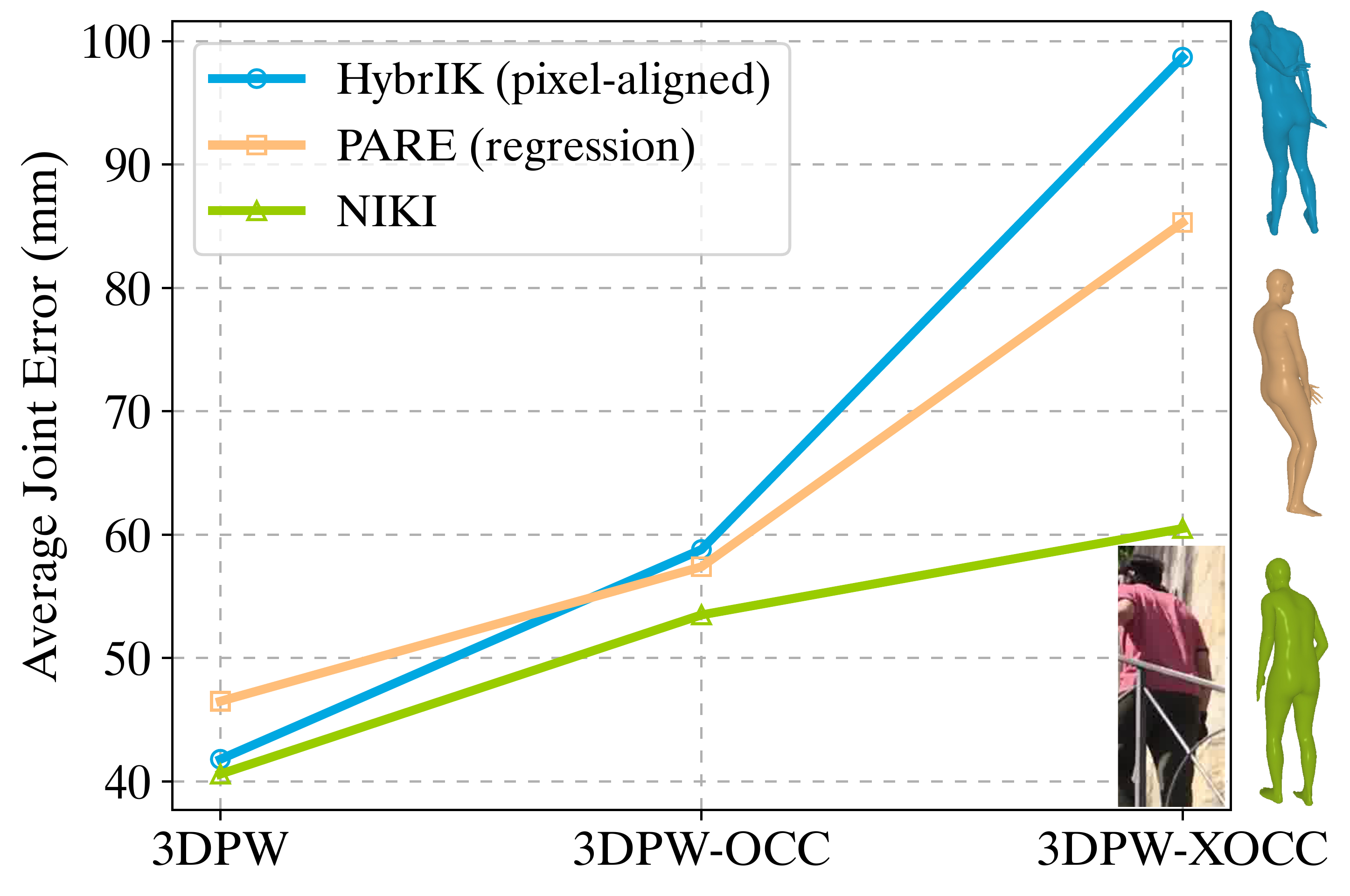}

   \caption{\textbf{Trade-off between pixel-aligned accuracy and robustness.} From 3DPW to 3DPW-XOCC, the degree of occlusion increases. The pixel-aligned approach performs well only in non-occlusion cases. The direct regression approach is more robust to occlusions but less accurate in non-occlusion cases. NIKI shows high accuracy and strong robustness simultaneously. Illustrative results on the 3DPW-XOCC dataset are shown on the right.}
   \label{fig:intro}
\end{figure}

Existing state-of-the-art approaches rely on pixel-aligned local evidence, \eg, 3D keypoints~\cite{li2021hybrik,iqbal2021kama}, mesh vertices~\cite{moon2020i2l}, and mesh-aligned features~\cite{zhang2021pymaf}, to perform accurate human pose and shape estimation. Although the local evidence helps obtain high accuracy in standard benchmarks, it fails when the mesh-image correspondences are unavailable due to occlusions and truncations. These pixel-aligned approaches sacrifice robustness to occlusions for high accuracy in non-occlusion scenarios. On the other hand, direct regression approaches are more robust to occlusions. Such approaches directly predict a set of pose and shape parameters with neural networks. By encoding human body priors in the networks, they predict a more physiologically plausible result than the pixel-aligned approaches in severely occluded scenarios. However, direct regression approaches use all pixels to predict human pose and shape, which is a highly non-linear mapping and suffers from image-mesh misalignment. Recent work~\cite{kocabas2021pare,khirodkar2022occluded} adopts guided attention to leverage local evidence for better alignment. Nevertheless, in non-occlusion scenarios, direct regression approaches are still not as accurate as the pixel-aligned approaches that explicitly model the local evidence. Fig.~\ref{fig:intro} shows the performance of the state-of-the-art pixel-aligned and regression approaches in scenarios with different levels of occlusions. These two types of approaches cannot achieve mesh-image alignment and robustness at the same time.
% which makes them sensitive to small perturbations of the input and suffer from image-mesh misalignment.

% In this work, we propose a new approach, NIKI (\textit{\textbf{N}eural \textbf{I}nverse \textbf{K}inematics algorithm with \textbf{I}nvertible neural networks}), to improve robustness to occlusions while maintaining pixel-aligned accuracy.

In this work, we propose NIKI, a Neural Inverse Kinematics (IK) algorithm with Invertible neural networks, to improve robustness to occlusions while maintaining pixel-aligned accuracy. IK algorithms are widely adopted in pixel-aligned approaches~\cite{li2021hybrik,iqbal2021kama} to obtain mesh-image alignment in non-occlusion scenarios. However, existing IK algorithms only focus on estimating the body part rotations that best explain the joint positions but do not consider the plausibility of the estimated poses.
% Therefore, they cannot correct the implausible joint positions, and the output human pose inherits the joint errors when the body joints are unreliable due to occlusions.
Therefore, the output human pose inherits the errors from joint position estimation, which is especially severe in occlusion scenarios.
% Previous inverse kinematics (IK) algorithms~\cite{li2021hybrik,voleti2022smpl} only focus on estimating the body part rotations that best explain the joint positions but do not consider the plausibility of the estimated human poses.
% % Although they obtain mesh-image alignment in non-occlusion cases, when the body joints are unreliable due to occlusions, they cannot correct the implausible joint positions, and the output human pose inherits the joint errors.
% Although they can obtain mesh-image alignment in non-occlusion cases, they cannot correct the implausible joint positions, and the output human pose inherits the joint errors when the body joints are unreliable due to occlusions.
In contrast, NIKI is robust to unreliable joint positions by modeling the bi-directional pose error. We build the bijective mapping between the Euclidean joint position space and the combined space of the 3D joint rotation and the latent error. The latent error indicates how the joint positions deviate from the manifold of plausible human poses. The output rotations are robust to erroneous joint positions since we have explicitly removed the error information by supervising the output marginal distribution in the inverse direction.
% Such error information is explicitly decoupled by supervising the output marginal distribution in the inverse direction.
% The output rotations are robust to erroneous joint positions by removing the error information.
In the forward direction, we introduce the zero-error boundary conditions, which enforce the solved rotations to explain the reliable joint positions and improve mesh-image alignment. The invertible neural network (INN) is trained in both forward and inverse directions.
% The solved rotations can accurately explain the reliable joint positions by additionally training on the well-defined forward kinematics (FK) process.
Since forward kinematics (FK) is deterministic and easy to understand, it aids the INN in learning the complex IK process through inherent bijective mapping. To further improve the interpretability of the IK network, we emulate the analytical IK algorithm by decomposing the complete rotation into the twist rotation and the swing-dependent joint position with two consecutive invertible networks.
% The first INN maps the joint positions to the swing rotations, and the second INN maps the twist and swing rotations to the complete body part rotations.
% Additionally, we show that INNs are flexible and can build bijective mapping in the time domain. We extend NIKI to incorporate temporal information for smooth human motion predictions.

We benchmark NIKI on {3DPW}~\cite{3dpw}, {AGORA}~\cite{patel2021agora}, {3DOH}~\cite{zhang2020object}, {3DPW-OCC}~\cite{3dpw}, and our proposed {3DPW-XOCC} datasets. {3DPW-XOCC} is augmented from the original {3DPW} dataset with extremely challenging occlusions and truncations. NIKI shows robust reconstructions while maintaining pixel-aligned accuracy, demonstrating state-of-the-art performance in both occlusions and non-occlusion benchmarks. The main contributions of this paper are summarized as follows:
\begin{itemize}
   \item We present a framework with a novel error-aware inverse kinematics algorithm that is robust to occlusions while maintaining pixel-aligned accuracy.
   \item We propose to decouple the error information from plausible human poses by learning a pose-independent error embedding in the inverse process and enforcing zero-error boundary conditions during the forward process using invertible neural networks.
   \item Our approach outperforms previous pixel-aligned and direct regression approaches on both occlusions and non-occlusion benchmarks.
\end{itemize}

\section{Related Work}

\paragraph{3D Human Pose and Shape Estimation.}
Prior work estimates 3D human pose and shape by outputting the parameters of statistical human body models~\cite{anguelov2005scape,loper2015smpl,pavlakos2019expressive,xu2020ghum,osmansupr}. Initial work follows the optimization paradigm~\cite{bogo2016keep,pavlakos2019expressive,lassner2017unite,varol2018bodynet}. SMPLify~\cite{bogo2016keep} is the first automated approach that fits SMPL parameters to 2D keypoint observations. This paradigm is further extended to silhouette~\cite{lassner2017unite} and volumetric grids~\cite{varol2018bodynet}.

Recently, learning-based paradigms have gained much attention with the advances in deep neural networks. Existing work can be categorized into two classes: direct regression approaches and pixel-aligned approaches. Direct regression approaches use deep neural networks to regress the pose and shape parameters directly~\cite{hmr,kanazawa2019learning,spin,vibe,wan2021encoder,kocabas2021pare,kocabas2021spec,li2022d}. Intermediate representations are used as the weak supervision to improve the regression performance, \eg, 2D keypoints~\cite{hmr} and body/part segmentation~\cite{pavlakos2018learning}. Several studies~\cite{spin,joo2021exemplar} leverage the optimization paradigm to introduce the pseudo ground truth for better supervision. Pixel-aligned approaches explicitly exploit pixel-aligned local evidence to estimate the pose and shape parameters. Moon \etal~\cite{moon2020i2l} use the vertex positions to regress the SMPL parameters. Li \etal~\cite{li2021hybrik} and Iqbal \etal~\cite{iqbal2021kama} propose to map the 3D keypoints to pose parameters. Zhang \etal~\cite{zhang2021pymaf} propose the mesh-aligned feedback loop to predict the aligned SMPL parameters. Explicitly modeling local evidence contributes to the state-of-the-art performance of pixel-aligned approaches.

Although pixel-aligned approaches achieve high accuracy in standard benchmarks, they are vulnerable to occlusions and truncations. When the local evidence is not reliable or even does not exist in occluded and truncated cases, such approaches predict physiologically implausible results. Direct regression approaches~\cite{jiang2020coherent,sun2021monocular,zhang2020object,kocabas2021pare,khirodkar2022occluded} are more robust to occlusions and truncations but less accurate in non-occlusion scenarios.
% Jiang \etal~\cite{jiang2020coherent} use an interpenetration loss to avoid collision and an ordinal loss to improve depth coherence.
% Sun \etal~\cite{sun2021monocular} construct a repulsion field to improve the robustness to person-person occlusions.
Zhang \etal~\cite{zhang2020object} use the saliency map to infer object-occluded human bodies. Kocabas \etal~\cite{kocabas2021pare} propose part-guided attention to exploit the information about the visibility of body parts. Khirodkar \etal~\cite{khirodkar2022occluded} use body centermaps to exploit the spatial context. A number of studies~\cite{pavlakos2019expressive,rempe2021humor,tiwari2022pose} propose to use pose prior to improve the plausibility of the estimated poses. Although the local evidence is implicitly used in recent regression approaches, pixel-aligned approaches still dominate non-occlusion benchmarks.

In this work, we combine the merits of pixel-aligned approaches and direct regression approaches. NIKI maintains pixel-aligned accuracy by aligning with the body joints via inverse kinematics while achieving robustness to occlusions and truncations with bi-directional error decoupling.

\paragraph{Inverse Kinematics.} The inverse kinematics (IK) process finds the relative rotations to produce the desired positions of body joints. It is an ill-posed problem because of the information loss in the forward process. Traditional numerical approaches~\cite{balestrino1984robust,wolovich1984computational,girard1985computational,klein1983review,wampler1986manipulator,buss2005selectively} are time-consuming due to iterative optimization. The heuristic approaches such as CDC~\cite{luenberger1984linear}, FABRIK~\cite{aristidou2011fabrik}, and IK-FA~\cite{rokbani2015ik} are more efficient and have a lower computation cost for each heuristic iteration. Recent work~\cite{csiszar2017solving,villegas2018neural} has started using neural networks to solve the IK problem.
% Csiszar \etal~\cite{csiszar2017solving} use a three-layer MLP to predict the rotations per joint. Villegas \etal~\cite{villegas2018neural} use an RNN model to directly regress the rotations.
Zhou \etal~\cite{zhou2019continuity} train a four-layer MLP network to predict the 3D human pose parameterized as 6D vectors. Li \etal~\cite{li2021hybrik} propose a hybrid analytical-neural solution to accurately predict the body part rotations. Oreshkin \etal~\cite{oreshkin2021protores} propose to use prototype encoding to predict rotations from sparse user inputs. Voleti \etal~\cite{voleti2022smpl} extend the same model to arbitrary skeletons. The work of Ardizzone \etal~\cite{ardizzone2018analyzing} is most related to us. They use invertible neural networks (INNs) to solve inverse problems, including the toy inverse kinematics problem in 2D space. However, similar to all the aforementioned approaches, they assume the input body joints are reliable, resulting in vulnerability to occlusions and truncations.

% The joint positions are given precisely, and the kinematics chain is a simple three-joint arm.

% All the aforementioned approaches assume the input body joints are reliable, resulting in the vulnerability to the occluded and truncated scenes.

\paragraph{Invertible Neural Network in HPS Estimation.}
Modeling the conditional posterior of an inverse process is a classical statistical task. Wehrbein \etal~\cite{wehrbein2021probabilistic} propose to estimate 3D human poses from 2D poses by capturing lost information with INNs. Several studies~\cite{xu2020ghum,zanfir2020weakly} leverage INNs to build priors for 3D human pose estimation. The pose priors are learned by normalizing flows that are built with INNs.
% Normalizing flows are generative models which produce tractable distributions with exact sampling and density evaluation through INNs.
% Xu \etal~\cite{xu2020ghum} propose novel 3D human models with the kinematic prior based on normalizing flows. Zanfir \etal~\cite{zanfir2020weakly} use normalizing flows to build kinematic priors.
Biggs \etal~\cite{biggs20203d} propose to use the learned prior from normalizing flows to resolve ambiguities. Kolotouros \etal~\cite{kolotouros2021probabilistic} propose a conditional distribution with normalizing flows as a function of the input to combine information from different sources. Li \etal~\cite{li2021human} leverage normalizing flows to capture the underlying residual log-likelihood of the output and propose a novel regression paradigm from the perspective of maximum likelihood estimation. Unlike previous methods, our approach leverages the property of bijective mapping in INNs to decouple joint errors and solve the inverse kinematics problem robustly.

\section{Method}

In this section, we present NIKI, a neural inverse kinematics solution for 3D human pose and shape estimation. We first review the formulation of existing analytical and MLP-based IK algorithms in \S\ref{sec:ik}. In \S\ref{sec:ik_inn}, we introduce the proposed INN-based IK algorithm with bi-directional error decoupling. In \S\ref{sec:train}, we present the overall human pose estimation framework and the learning objective. Then we elaborate on the proposed IK-specific invertible architecture in \S\ref{sec:arch}.
% In \S\ref{sec:arch}, we elaborate on the proposed IK-specific invertible architecture.
% Then we present the overall pose estimation framework and the training losses in \S\ref{sec:train}.
Finally, we provide the necessary implementation details in \S\ref{sec:implement}

\subsection{Preliminaries}
\label{sec:ik}
% 1. ik, fk; 2. inn; 3. information perserved, error decoupling;

% We study the problem of inverse kinematics (IK) using invertible neural networks (INNs).
The IK process is to find the corresponding body part rotations that explain the input body joint positions, while the forward kinematics (FK) process computes the desired joint positions based on the input rotations. The FK process is well-defined, but the transformation from joint rotations to joint positions incurs an information loss, \ie, multiple rotations could correspond to one position, resulting in an ill-posed IK process. Here, we follow HybrIK~\cite{li2021hybrik} to consider twist rotations for information integrity.

\begin{figure}[tbp]
    \centering
    \includegraphics[width=0.96\linewidth]{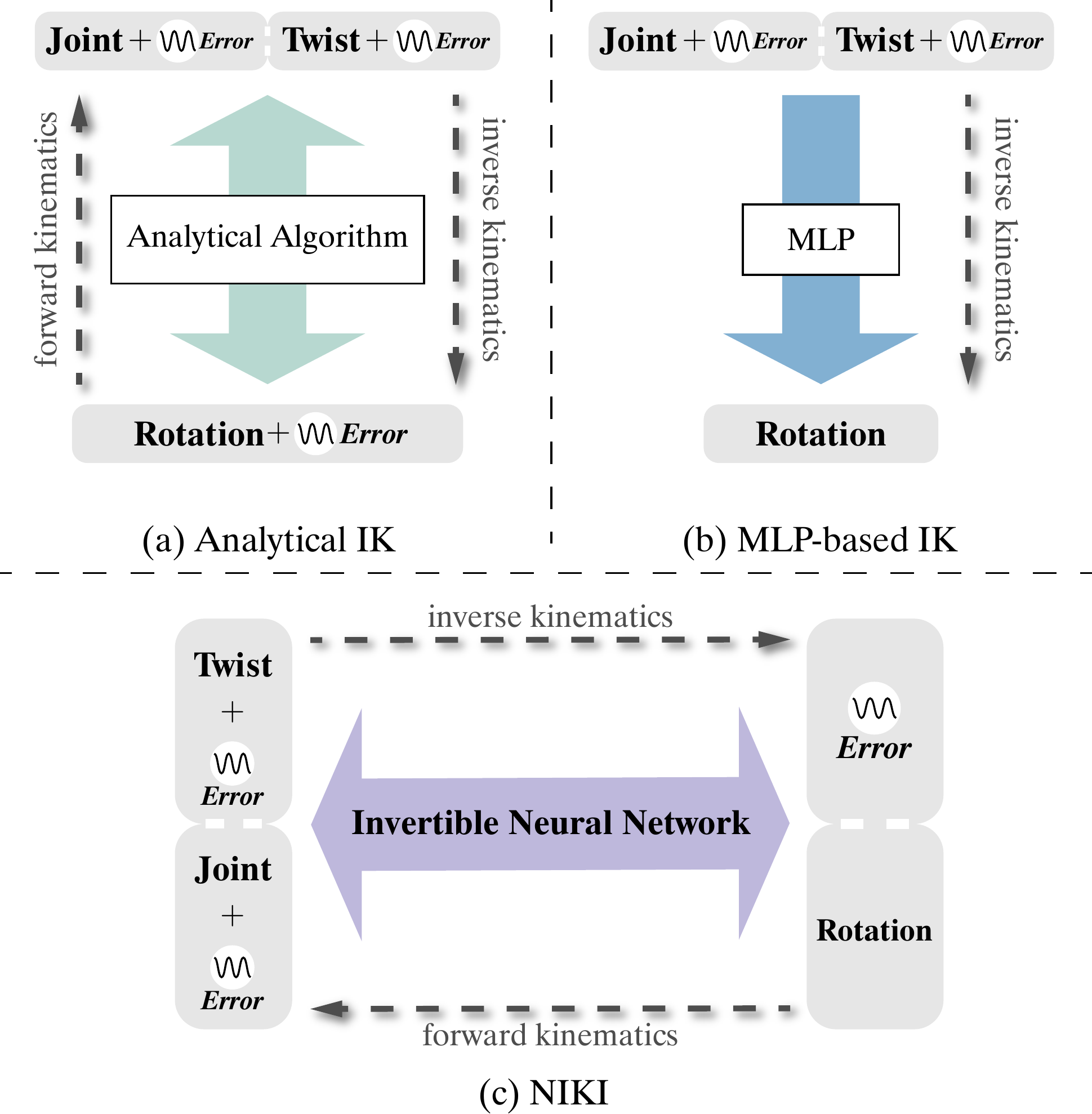}
 
    \caption{\textbf{Illustration} of (a) analytical IK, (b) feedforward MLP-based IK, and (c) NIKI with bi-directional error decoupling.}
    \label{fig:decouple}
\end{figure}

The conventional IK algorithms only require the output rotations to match the input joint positions but ignore the errors of the joint positions and the plausibility of the body pose. Therefore, the errors of the joint positions will be accumulated in the joint rotations (Fig.~\ref{fig:decouple}a). This process can be formulated as:
\begin{equation}
    \underbrace{\mathbf{R} + \epsilon_r}_{\text{erroneous output}} = \text{IK}_{\text{Analytical}}(\underbrace{\mathbf{p} + \epsilon_p, \boldsymbol{\phi} + \epsilon_{\phi}}_{\text{erroneous input}} ~|~ \boldsymbol{\beta}),
\end{equation}
where $\mathbf{R}$ denotes the underlying plausible rotations, $\epsilon_r$ denotes the accumulated error in estimated rotations, $\mathbf{p}$ denotes the underlying plausible joint positions, $\epsilon_p$ denotes the position errors, $\boldsymbol{\phi}$ denotes the underlying plausible twist rotations, $\epsilon_{\phi}$ denotes the twist error, and $\boldsymbol{\beta}$ denotes the body shape parameters.

A straightforward solution to improving the robustness of the IK algorithms is using the standard regression model~\cite{csiszar2017solving,zhang2020object} to approximate the underlying plausible rotations $\mathbf{R}$ given the erroneous input (Fig.~\ref{fig:decouple}b):
% To improve the robustness of the IK process to joint position and twist rotation errors, a straightforward solution is to train a standard regression model~\cite{csiszar2017solving,zhang2020object} to output the underlying plausible rotations given deviated input (Fig.~\ref{fig:decouple}b):
\begin{equation}
    \mathbf{R} \approx \text{IK}_{\text{MLP}}(\mathbf{p} + \epsilon_p, \boldsymbol{\phi} + \epsilon_{\phi} ~|~ \boldsymbol{\beta}).
\end{equation}
% As shown in Fig.~\ref{fig:decouple}(b), the model can be trained to predict the correct rotations with erroneous input.  
Indeed, modeling the IK process with classical neural networks, \eg, MLP, can improve the robustness. However, the output rotations are less sensitive to the change of the joint positions. The errors are highly coupled with the joint positions. Without explicitly decoupling errors from plausible human poses, it is difficult for the network to distinguish between reasonable and abnormal changes in joint positions. Therefore, the output rotations cannot accurately track the movement of the body joints. In practice, we find that the feedforward neural networks could improve performance in occlusion cases but cause performance degradation in non-occlusion cases, where accurate mesh-image alignment is required. Detailed comparisons are provided in Tab.~\ref{table:ablation}.

\begin{figure}[t]
    \centering
    \includegraphics[width=\linewidth]{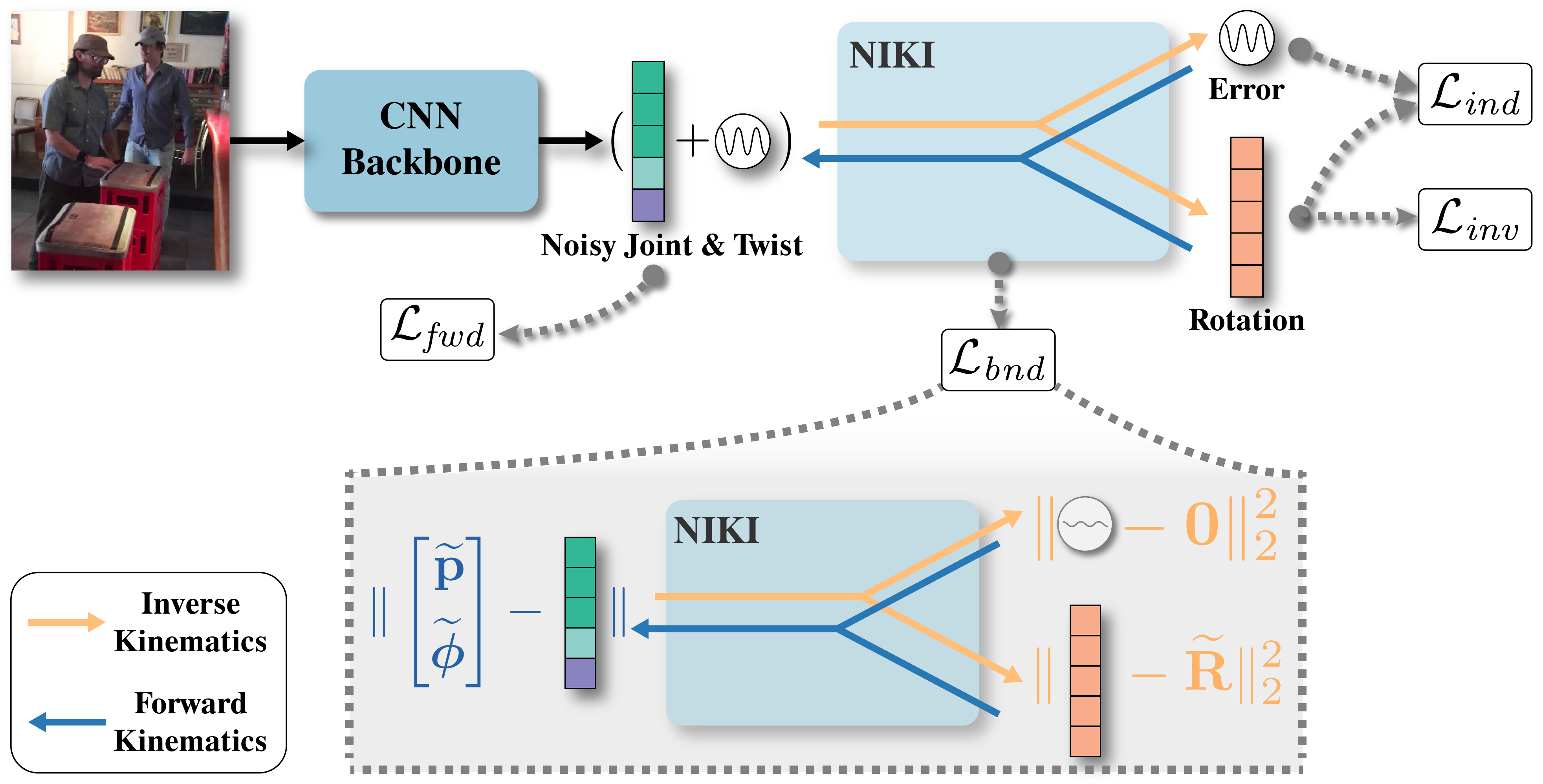}
 
    \caption{\textbf{Overview of the proposed framework.} The input image is fed into the CNN backbone network to estimate the initial joint positions and twist rotations, followed by NIKI to solve the joint rotations.}
    \label{fig:framework}
 \end{figure}

\subsection{Inverse Kinematics with INNs}
\label{sec:ik_inn}
In this work, to improve the robustness of IK to occlusions while maintaining the sensitivity to non-occluded body joints, we propose to use the invertible neural network (INN) to model bi-directional errors explicitly (see Fig.~\ref{fig:decouple}c). In contrast to the conventional methodology, we learn the IK model $g(\cdot ; \boldsymbol{\beta}, \theta)$ jointly with the FK model $f(\cdot ; \boldsymbol{\beta}, \theta)$:
\begin{align}
    [\mathbf{p} + \epsilon_p, \boldsymbol{\phi} + \epsilon_{\phi}] &= f(\mathbf{R}, \mathbf{z}_r ; \boldsymbol{\beta}, \theta), \\
    [\mathbf{R}, \mathbf{z}_r] &= g(\mathbf{p} + \epsilon_p, \boldsymbol{\phi} + \epsilon_{\phi} ; \boldsymbol{\beta}, \theta),
\end{align}
where $\mathbf{z}_r$ is the error embedding that denotes how the input joint positions deviate from the manifold of plausible human poses. Notice that $f$ and $g$ share the same parameters $\theta$, and $f=g^{-1}$ is enforced by the invertible network architecture. We expect that simultaneously learning the FK and IK processes can benefit each other.
% An auxiliary training on the well-understood FK process can help the network learn the IK process.

% Thus simultaneously learning the FK and IK processes can benefit each other.

In the forward process, we can tune the error embedding $\mathbf{z}_r$ to control the error level of the body joint positions. The body part rotations will perfectly align with the joint positions by setting $\mathbf{z}_r$ to $0$, which means no deviation from the pose manifold. In the inverse process, the error is only reflected on $\mathbf{z}_r$. The rotation $\mathbf{R}$ keeps stable against the erroneous input.
% For example, if we set $\mathbf{z}_r$ to $0$, the body part rotations will perfectly align with the joint positions.
% The body part rotations can be kept sensitive to the movement of body joint positions by setting $\mathbf{z}_r$ to $0$, which means no deviation from the pose manifold and the rotations should perfectly align with the joint positions.

\paragraph{Decouple Error Information.}
% To explicitly model the error, we need to separate it from the pose information.
The input joints and twists to the IK process contain two parts of information: i) the underlying pose that lies on the manifold of plausible 3D human poses; ii) the error information that indicates how the input deviates from the manifold.
% We can obtain robust pose estimation by separating errors from the pose manifold.
We can obtain robust pose estimation by separating these two types of information.
Due to the bijective mapping enforced by the INN, all the input information is preserved in the output, and no new information is introduced. Therefore, we only need to remove the pose information from the output vector $\mathbf{z}_r$ in the inverse process. The vector $\mathbf{z}_r$ will automatically encode the remaining error information. To this end, we enforce the model to follow the independence constraint, which encourages $\mathbf{R}$ and $\mathbf{z}_r$ to be independent upon convergence, \ie, $p(\mathbf{z}_r | \mathbf{R}) = p(\mathbf{z}_r)$.
% Concretely, we enforce $\mathbf{R}$ and $\mathbf{z}_r$ to be independent upon convergence, \ie, $p(\mathbf{z}_r | \mathbf{R}) = p(\mathbf{z}_r)$, to remove the pose information from the vector $\mathbf{z}_r$.

After we separate the error information, we can manipulate the error embedding to let the model preserve the sensitivity to error-free body joint positions without compromising robustness. In particular, we constrain the error information in the forward process with the zero-error condition:
% This independence constraint allows us to model the error information in the inverse process.
% Once we separate the pose information and the error information,
% The independence constraint allows us to decouple the error information in the inverse process, which improves the robustness to erroneous joint positions. To further preserve the sensitivity to error-free body joint positions, we constrain the error information in the forward process with the zero-error condition:
% Although the independence constraint allows us to separate the error information from the pose information, the decoupling only happens in the inverse process. We need to further constrain the error information in the forward process to maintain sensitivity to error-free body joint positions. Here, we propose to enforce the model to satisfy the following zero-error condition:
% Besides, the model needs to satisfy the following zero-error conditions to maintain sensitivity to error-free body joint positions:
% To decouple the errors, the latent code $\mathbf{z}_r$ should have no pose information.
% we train the bijective mapping to satisfy the following the zero-error conditions:
% \begin{align}
%     [\mathbf{p}, \phi] &= f(\mathbf{R}, \bigzero ; \boldsymbol{\beta}, \theta), \\
%     [\mathbf{R}, \bigzero] &= g(\mathbf{p}, \boldsymbol{\phi} ; \boldsymbol{\beta}, \theta).
% \end{align}
\begin{equation}
    [\mathbf{p}, \phi] = f(\mathbf{R}, \bigzero ; \boldsymbol{\beta}, \theta).
\end{equation}
In this way, the rotations will track the joint positions and twist rotations accurately in non-occlusion scenarios. Besides, the zero-error condition can also be extended to the inverse process:
\begin{equation}
    [\mathbf{R}, \bigzero] = g(\mathbf{p}, \boldsymbol{\phi} ; \boldsymbol{\beta}, \theta).
\end{equation}

With the independence and zero-error constraints, the network is able to model the error information in both the forward and inverse processes, making NIKI robust to occlusions while maintaining pixel-aligned accuracy.
% With the constraints for the zero-error boundary, we explicitly let the model decouple the bi-directional errors.
% In the forward process, we can tune error embedding $\mathbf{z}_r$ to control the error level of the body joint positions. The body part rotations can keep sensitive to the movement of body joint positions by setting $\mathbf{z}_r$ to $0$. In the inverse process, the error decoupling makes the output rotations stable and robust to the erroneous input.

% The error decoupling in the inverse process makes the model robust to input errors, while the error modeling in the forward process keeps the body part rotations sensitive to the movement of error-free body joint positions.

\subsection{Decoupled Learning}
\label{sec:train}

The overview of our approach is illustrated in Fig.~\ref{fig:framework}. During inference, we first extract the joint positions and twist rotations with the CNN backbone, which are subsequently fed to the invertible network to predict the complete body part rotations. During training, we optimize FK and IK simultaneously in one network.
% During training, the invertible neural network allows us to simultaneously optimize FK and IK to train the same network.
Hereby, we perform FK and IK alternately with the additional independence loss and boundary loss. The gradients from both directions are accumulated before performing a parameter update.

\paragraph{Inverse Training.}
In the inverse iteration, the network predicts the body part rotations given the joint positions $\hat{\mathbf{p}}$ and twist rotations $\hat{\boldsymbol{\phi}}$ from the CNN backbone. The loss function is defined as:
\begin{equation}
    \mathcal{L}_{\textit{inv}} = \big\| \hat{\mathbf{R}}_{\textit{inv}} - \widetilde{\mathbf{R}} \big\|^2_2 + \big\| \text{FK}(\hat{\mathbf{R}}_{\textit{inv}}) - \text{FK}(\widetilde{\mathbf{R}}) \big\|_1,
    \label{eq:inv_r}
\end{equation}
with
\begin{equation}
    [\hat{\mathbf{R}}_{\textit{inv}}, \hat{\mathbf{z}}_r] = g(\hat{\mathbf{p}}, \hat{\boldsymbol{\phi}} ; \boldsymbol{\beta}, \theta),
\end{equation}
where $\widetilde{\mathbf{R}}$ represents the ground-truth rotations, and $\text{FK}(\cdot)$ denotes the analytical FK process to supervise the corresponding 3D joint positions of the predicted pose.

% For the twist-and-swing mapping model, we additionally supervise the swing rotations:
% \begin{equation}
%     \mathcal{L}^{\textit{inv}}_{\textit{sw}} = \big\| \hat{\mathbf{R}}^{\textit{inv}}_{\textit{sw}} - \widetilde{\mathbf{R}}_{\textit{sw}} \big\|^2_2,
% \end{equation}
% with
% \begin{equation}
%     [\hat{\mathbf{R}}^{\textit{inv}}_{\textit{sw}}, \hat{\mathbf{z}}_\textit{sw}] = g_1(\hat{\mathbf{p}} ; \boldsymbol{\beta}, \theta_1).
% \end{equation}

\paragraph{Forward Training.}
In the forward process, the network predicts the joint positions and twist rotations given the body part rotations. The error of the noisy predictions $\hat{\mathbf{p}}$ and $\hat{\boldsymbol{\phi}}$ should only be determined by the error embedding. Therefore, with the ground-truth rotations $\widetilde{\mathbf{R}}$ and error embedding $\hat{\mathbf{z}}_r$ obtained from the inverse iteration, the forward model should predict the same values as the CNN output:
\begin{equation}
    \mathcal{L}_{\textit{fwd}} = \| \hat{\mathbf{p}}_{\textit{fwd}} - \hat{\mathbf{p}} \|_1 + \| \hat{\boldsymbol{\phi}}_{\textit{fwd}} - \hat{\boldsymbol{\phi}} \|_2^2,
\end{equation}
with
\begin{equation}
    [\hat{\mathbf{p}}_{\textit{fwd}}, \hat{\boldsymbol{\phi}}_{\textit{fwd}}] = f(\widetilde{\mathbf{R}}, \hat{\mathbf{z}}_r ; \boldsymbol{\beta}, \theta).
\end{equation}

% For the twist-and-swing mapping model, we also supervise the swing rotations:
% \begin{equation}
%     \mathcal{L}^{\textit{fwd}}_{\textit{sw}} = \big\| \hat{\mathbf{R}}^{\textit{fwd}}_{\textit{sw}} - \hat{\mathbf{R}}^{\textit{inv}}_{\textit{sw}} \big\|^2_2,
% \end{equation}
% with
% \begin{equation}
%     [\hat{\mathbf{R}}^{\textit{fwd}}_{\textit{sw}}, \hat{\boldsymbol{\phi}}^{\textit{fwd}}] = f_2(\widetilde{\mathbf{R}}, \mathbf{z}_r ; \theta_2).
% \end{equation}

\paragraph{Independence Loss.}
% The characteristic of bijective mappings ensure that there is no information loss in both the forward and inverse transformations. When the input contains the pose information and the error information, the output must contain the same information. This is the basis of the error decoupling process.
% The latent error vector denotes how the joint positions deviate from the manifold of plausible human poses.
% which is difficult to define and learn in a supervised manner. Since the total amount of information is constant, we can enforce $\mathbf{z}_r$ to contain error information by supervising the inverse output $\mathbf{R}_{\textit{inv}}$ with pose rotations and
The latent error vector is learned in an unsupervised manner by making $\mathbf{R}_{\textit{inv}}$ and $\mathbf{z}_r$ independent of each other. The pose information in $\mathbf{R}_{\textit{inv}}$ is supervised by Eq.~\ref{eq:inv_r}. We then enforce the independence by penalizing the mismatch between the joint distribution of the rotations and error embedding $q\big(\hat{\mathbf{R}}_{\textit{inv}}, \mathbf{z}_r\big)$ and the product of marginal distributions $p\big(\widetilde{\mathbf{R}} \big)p(\mathbf{z})$:
\begin{equation}
    \mathcal{L}_{\textit{ind}} = \mathcal{D}(q\big(\mathbf{R}_{\textit{inv}}, \mathbf{z}_r\big), p\big(\widetilde{\mathbf{R}} \big)p(\mathbf{z})),
\end{equation}
where $\mathbf{z} \sim \mathcal{N}(0, \mathbf{I})$ follows the standard normal distribution, $\mathbf{I}$ is the identity matrix, and $\mathcal{D}(\cdot)$ denotes the Maximum Mean Discrepancy~\cite{gretton2012kernel}, which allows us to compare two probability distributions through samples.
In addition to the independence constraint, $\mathcal{L}_{\textit{ind}}$ encourages the error embedding $\mathbf{z}_r$ to follow the standard normal distribution $p(\mathbf{z})$, serving as a regularization.
% Follwing Ardizzone \etal~\cite{ardizzone2018analyzing}, we block the gradients of $\mathcal{L}_{\textit{ind}}$ with respective to $\hat{\mathbf{R}}_{\textit{inv}}$ to ensure the resulting updates only affect the predictions of $\mathbf{z}_r$ and do not worsen the prediction of rotations.

\paragraph{Boundary Condition Loss.}
To enforce the solved rotations to explain the reliable joint positions, we supervise the boundary cases where no error occurs. In the inverse process, the output error should be zero when the network is fed with the ground truth:
\begin{equation}
    \mathcal{L}^{\textit{i}}_{\textit{bnd}} = \| \hat{\epsilon}_r \|_2^2 + \| \hat{\mathbf{R}}_{\textit{bnd}} - \widetilde{\mathbf{R}} \|^2_2,
\end{equation}
with
\begin{equation}
    [\hat{\mathbf{R}}_{\textit{bnd}}, \hat{\epsilon}_r] = g(\widetilde{\mathbf{p}}, \widetilde{\boldsymbol{\phi}} ; \boldsymbol{\beta}, \theta),
\end{equation}
where $\widetilde{\mathbf{p}}$ and $\widetilde{\mathbf{\boldsymbol{\phi}}}$ denote the ground-truth joint positions and twist rotations, respectively.

% For the twist-and-swing mapping model, we also supervise the error embedding of the swing rotations.
% \begin{equation}
%     \mathcal{L}^{\textit{inv}}_{\textit{bnd},\textit{sw}} = \| \epsilon_{\textit{sw}} \|_2^2, 
% \end{equation}
% with
% \begin{equation}
%     [\mathbf{R}_{\textit{bnd}}^{\textit{sw}}, \epsilon_{\textit{sw}}] = g_1(\widetilde{\mathbf{p}}, \widetilde{\boldsymbol{\phi}} ; \boldsymbol{\beta}, \theta_1).
% \end{equation}
In the forward process, the joint positions and twist rotations should map to the ground truth when the input error vector $\mathbf{z}_r$ is $\mathbf{0}$:
\begin{equation}
    \mathcal{L}^{\textit{f}}_{\textit{bnd}} = \| \hat{\mathbf{p}}_{\textit{bnd}} - \widetilde{\mathbf{p}} \|_1 + \| \hat{\boldsymbol{\phi}}_{\textit{bnd}} - \widetilde{\boldsymbol{\phi}} \|_2^2,
\end{equation}
with
\begin{equation}
    [\hat{\mathbf{p}}_{\textit{bnd}}, \hat{\boldsymbol{\phi}}_{\textit{bnd}}] = f(\widetilde{\mathbf{R}}, \mathbf{0} ; \boldsymbol{\beta}, \theta).
\end{equation}

Overall, the total loss of NIKI is:
\begin{equation}
\begin{aligned}
    \mathcal{L} = &\lambda_{\textit{inv}} \mathcal{L}_{\textit{inv}} + \lambda_{\textit{fwd}} \mathcal{L}_{\textit{fwd}} \\ + &\lambda_{\textit{ind}} \mathcal{L}_{\textit{ind}} + \lambda_{\textit{bnd}}^\textit{i} \mathcal{L}_{\textit{bnd}}^\textit{i} + \lambda_{\textit{bnd}}^\textit{f} \mathcal{L}_{\textit{bnd}}^\textit{f},
\end{aligned}
\end{equation}
where $\lambda_{\textit{inv}}, \lambda_{\textit{fwd}}, \lambda_{\textit{ind}}, \lambda_{\textit{bnd}}^\textit{i}, \lambda_{\textit{bnd}}^\textit{f}$ are the scalar coefficients to balance the loss terms.

\subsection{Invertible Architecture}
\label{sec:arch}

\paragraph{One-Stage Mapping.}
To build a fully invertible neural network for inverse kinematics, we build the one-stage mapping model using RealNVP~\cite{dinh2016density}. Since the IK and FK processes require the skeleton template, we extend the INN to incorporate the conditional shape parameters input. The basic block of the network contains two reversible coupling layers conditioned on the shape parameters.
% The input vector $\mathbf{u}$ of the block is split into two parts, $\mathbf{u}_1$ and $\mathbf{u}_2$, which are subsequently transformed with coefficients $\exp(s_i)$ and $t_i$ ($i \in \{1, 2\}$) by the two affine coupling layers:
% \begin{align}
%     \mathbf{v}_1 = \mathbf{u}_1 \odot \exp(s_2(\mathbf{u}_2, \boldsymbol{\beta})) + t_2(\mathbf{u}_2, \boldsymbol{\beta}), \label{eq:affine1}\\
%     \mathbf{v}_2 = \mathbf{u}_2 \odot \exp(s_1(\mathbf{v}_1, \boldsymbol{\beta})) + t_1(\mathbf{v}_1, \boldsymbol{\beta}), \label{eq:affine2}
% \end{align}
% where $\mathbf{v} = [\mathbf{v}_1, \mathbf{v}_2]$ is the output vector of the block and $\odot$ denotes element-wise multiplication. The coefficients of the affine transformation can be learned by arbitrarily complex functions, which do not need to be invertible. The invertibility is guaranteed by the affine transformation in Eq.~\ref{eq:affine1} and \ref{eq:affine2}.
The overall network consists of multiple blocks connected in series to increase capacity.
% To increase capacity, the overall network consists of multiple block connect in stack.
Besides, since the invertible network requires the input and output vectors to have the same dimension, we follow previous work~\cite{ardizzone2018analyzing} and pad zeros to the input.

% \begin{figure}[t]
%     \centering
%     % \fbox{\rule{0pt}{2in} \rule{0.9\linewidth}{0pt}}
%     \includegraphics[width=0.8\linewidth]{figures/arch.pdf}
 
%     \caption{\textbf{Twist-and-Swing mapping.}}
%     \label{fig:arch}
% \end{figure}

\paragraph{Twist-and-Swing Mapping.}
% Although we can model both the FK and IK processes at the same time by treating the invertible neural network as a black box,
Although treating the invertible neural network as a black box can let us model both the FK and IK processes at the same time, we further emulate the analytical IK algorithm to improve the performance and interpretability. Specifically, we follow the twist-and-swing decomposition~\cite{li2021hybrik} and divide the IK process into two steps: i) from joint positions to swing rotations;  ii) from twist and swing rotations to complete rotations. The two-step mapping is implemented by two separate invertible networks:
\begin{align}
    [\mathbf{R}_{\textit{sw}}, \mathbf{z}_{\textit{sw}}] &= g_1(\mathbf{p} + \epsilon_p ; \boldsymbol{\beta}, \theta_1), \\
    [\mathbf{R}, \mathbf{z}_r] &= g_2(\mathbf{R}_{\textit{sw}} ,\boldsymbol{\phi} + \epsilon_{\phi} ; \theta_2),
\end{align}
where $\mathbf{R}_{\textit{sw}}$ is the swing rotations, and $\mathbf{z}_{\textit{sw}}$ indicates the deviation from the plausible swing rotation manifold.
% The first step uses a conditional invertible network, which takes as input the noisy joint positions $\mathbf{p} + \epsilon_p$ with the shape condition $\boldsymbol{\beta}$ and predict the swing rotations $\mathbf{R}_{\textit{sw}}$ and the error of swing rotations $\mathbf{z}_{\textit{sw}}$. The second step uses a invertible network without conditions.
% It maps the swing rotations $\mathbf{R}_{\textit{sw}}$ and the twist rotations $\boldsymbol{\phi}$ to the complete rotations $\mathbf{R}$ and the latent code $\mathbf{z}_{\text{r}}$ that denotes the error of the complete rotations, which further includes the implausible twisting information.
% The whole process of the twist-and-swing mapping is illustrated in Fig.~\ref{fig:arch}(b).

% The two-step twist-and-swing mapping improves the interpretability of the IK network.
Since the mappings are bijective, the FK process also follows the twist-and-swing procedure but inversely. We have $f = f_1 \circ f_2 = g_1^{-1} \circ g_2^{-1} = g^{-1}$. In the FK process, the body part rotations are first decomposed into twist and swing rotations. Then the swing rotations are transformed into the joint positions. The intermediate supervision of swing rotations is used in both the forward and inverse training.

\paragraph{Temporal Extension.}
The invertible framework is flexible.
% It can be easily extended to temporal input and incorporate motion information to solve the IK problem.
It can be easily extended to solving the IK problem with temporal inputs.
% by incorporating motion information.
The model with static inputs can only identify the errors related to physiological implausibility. In contrast, the temporal model further improves motion smoothness by decoupling errors of implausible human body movements. More details are provided in the supplementary material.
% (TODO: model architecture)

\subsection{Implementation Details.}
\label{sec:implement}

We adopt HybrIK~\cite{li2021hybrik} as the CNN backbone to predict the noisy body joint positions and the twist rotations. The input of the IK model includes the joint positions $\mathbf{p} \in \mathbb{R}^{3K}$, twist rotations parameterized in 2-dimensional vectors, \ie, $\boldsymbol{\phi} \in \mathbb{R}^{2(K - 1)}$, and the confidence scores of each joint $\mathbf{s} \in \mathbb{R}^{K}$, where $K$ denotes the total number of human body joints.
% The total valid dimension of the input is $D_{\textit{in}} = 6K - 2$.
The output of the model consists of the body part rotations parameterized as a 6D vector for each part, \ie, $\mathbf{R} \in \mathbb{R}^{6K}$, and the error embedding $\mathbf{z}_r \in \mathbb{R}^{D_z}$.
% Therefore, the valid output dimension is $D_{\textit{out}} = 6K + D_z$. Since $D_{\textit{out}} > D_{\textit{in}}$,
The IK model is conditioned on the shape parameters $\boldsymbol{\beta}$, which is also predicted by the CNN backbone.
% For the twist-and-swing mapping model, only the first INN is conditioned on the shape parameters.
We pad the input with a zero vector with the dimension $M = D_z + 2$ to satisfy the dimension constraint of the invertible neural network.
%%%
% The input of the IK model includes the joint positions, twist rotations parameterized in 2-dimensional vectors for each joint, and the confidence scores of each joint.
% % The total valid dimension of the input is $D_{\textit{in}} = 6K - 2$.
% The output of the model consists of the body part rotations parameterized as 6D vectors and the error embedding.
%%%
% Therefore, the valid output dimension is $D_{\textit{out}} = 6K + D_z$. Since $D_{\textit{out}} > D_{\textit{in}}$,
% We pad the input with a zero vector to satisfy the dimension constraints of the invertible neural network.
The networks are trained with the Adam solver for 50 epochs with a mini-batch size of $64$. The learning rate is set to $1 \times 10^{-3}$ at first and reduced by a factor of 10 at the 30th and 40th epochs. Implementation is in PyTorch. Detailed architectures are provided in the supplementary material.

\section{Experiments}

\paragraph{Datasets.}
We employ the following datasets in our experiments:
(1) \texttt{3DPW}~\cite{3dpw}, an outdoor benchmark for 3D human pose estimation.
% It contains $60$ video sequences obtained from a hand-held moving camera.
% The ground-truth poses are obtained via wearable IMU sensors.
% (2) \texttt{Human3.6M}~\cite{h36m}, an indoor benchmark for 3D human pose estimation. We use $5$ subjects (S1, S5, S6, S7, S8) for training and $2$ subjects (S9, S11) for testing.
(2) \texttt{AGORA}~\cite{patel2021agora}, a synthetic dataset with challenging environmental occlusions.
(3) \texttt{3DOH}~\cite{zhang2020object}, a 3D human dataset where human activities are occluded by objects.
(4) \texttt{3DPW-OCC}~\cite{3dpw}, a different split of the original 3DPW dataset with a occluded test set.
(5) \texttt{3DPW-XOCC}, a new benchmark for 3D human pose estimation with extremely challenging occlusions and truncations. We simulate occlusions and truncations by randomly pasting occlusion patches and cropping frames with truncated windows.

\paragraph{Training and Evaluation.} NIKI is trained on \texttt{3DPW}~\cite{3dpw} and \texttt{Human3.6M}~\cite{h36m} and evaluated on \texttt{3DPW}~\cite{3dpw} and \texttt{3DPW-XOCC}~\cite{3dpw} to benchmark the performance on both occlusions and non-occlusion scenarios. We use the \texttt{AGORA}~\cite{patel2021agora} train set only when conducting experiments on its test set. For evaluations on the \texttt{3DPW-OCC}~\cite{3dpw} and \texttt{3DOH}~\cite{zhang2020object} datasets, we train NIKI on \texttt{COCO}~\cite{coco}, \texttt{Human3.6M}~\cite{h36m}, and \texttt{3DOH}~\cite{zhang2020object} for a fair comparison. Procrustes-aligned mean per joint position error (PA-MPJPE) and mean per joint position error (MPJPE) are reported to assess the 3D pose accuracy. Per vertex error (PVE) is also reported to evaluate the estimated body mesh.
% For \texttt{3DPW}, \texttt{3DPW-OCC} and \texttt{3DPW-XOCC}, we also report per vertex error (PVE).

\begin{table}[t]
    \begin{center}
        \resizebox{\linewidth}{!}
        {
            \begin{tabular}{ll|ccc}

            \toprule
			& & \multicolumn{3}{c}{ \texttt{3DPW} }  \\
			\cmidrule(lr){3-5}
            & Method & ~MPJPE~$\downarrow$~ & ~PA-MPJPE~$\downarrow$~ & ~PVE~$\downarrow$~ \\
            \midrule
            \parbox[t]{2mm}{\multirow{5}{*}{\rotatebox[origin=c]{90}{Regression}}} & HMR~\cite{hmr} & 130.0 & 81.3 & - \\
            & SPIN~\cite{spin} & 96.9 & 59.2 & 116.4 \\
            & ROMP~\cite{sun2021monocular} & 85.5 & 53.3 & 103.1 \\
            & METRO~\cite{lin2021end} & 77.1 & 47.9 & 88.2 \\
            & PARE~\cite{kocabas2021pare} & 74.5 & 46.5 & 88.6 \\
            \midrule
            \parbox[t]{2mm}{\multirow{6}{*}{\rotatebox[origin=c]{90}{Pixel-aligned}}} & PyMAF~\cite{zhang2021pymaf} & 92.8 & 58.9 & 110.1  \\
            & I2L~\cite{moon2020i2l} & 93.2 & 58.6 & - \\
            & KAMA~\cite{iqbal2021kama} & - & 51.1 & 97.0 \\
            & Mesh Graphormer~\cite{lin2021mesh} & 74.7 & 45.6 & 87.7 \\
            & HybrIK (ResNet)~\cite{li2021hybrik} & 76.2 & 45.1 & 89.1 \\
            & HybrIK (HRNet)~\cite{li2021hybrik} & 72.9 & 41.8 & 88.6 \\
			% \cmidrule(lr){2-5}
            \midrule
            & NIKI (One-Stage) & {71.7} & {41.0} & 86.9 \\
            & NIKI (Twist-and-Swing) & \textbf{71.3} & \textbf{40.6} & \textbf{86.6} \\
            \bottomrule
            \end{tabular}
        }
        \caption{\textbf{Quantitative comparisons with state-of-the-art methods on the \texttt{3DPW} dataset.} Symbol ``-'' means results are not available.}
        \label{table:3dpw}
    \end{center}
\end{table}

\begin{table}[t]
    \begin{center}
        \resizebox{\linewidth}{!}
        {
            \begin{tabular}{l|ccc}

            \toprule
			 & \multicolumn{3}{c}{ \texttt{3DPW-XOCC} }  \\
			\cmidrule(lr){2-4}
            Method & ~MPJPE~$\downarrow$~ & ~PA-MPJPE~$\downarrow$~ & ~PVE~$\downarrow$~ \\
            \midrule
            HybrIK~\cite{li2021hybrik} & 148.3 & 98.7 & 164.5 \\
            PARE~\cite{kocabas2021pare} & 139.4 & 85.3 & 151.6 \\
            PARE$^*$~\cite{kocabas2021pare} & 114.2 & 67.7 & 133.0 \\
            \midrule
            NIKI (One-Stage) & {117.0} & {64.4} & 135.6 \\
            NIKI (Twist-and-Swing) & \textbf{110.7} & \textbf{60.5} & \textbf{128.6} \\

            \bottomrule
            \end{tabular}
        }
        \caption{\textbf{Quantitative comparisons with state-of-the-art methods on the \texttt{3DPW-XOCC} dataset.} Symbol $*$ means finetuning on the \texttt{3DPW-XOCC} train set.}
        \label{table:3dpw_xocc}
    \end{center}
\end{table}

\subsection{Comparison to State-of-the-art Methods}

We evaluate NIKI on both standard and occlusion-specific benchmarks. Tab.~\ref{table:3dpw} compares NIKI with previous state-of-the-art HPS methods on the standard \texttt{3DPW} dataset. We report the results of NIKI with one-stage mapping and twist-and-swing mapping. We can observe that in the standard benchmark, pixel-aligned approaches obtain better performance than direct regression approaches. NIKI significantly outperforms the most accurate direct regression approach by 5.9 mm on PA-MPJPE (12.7\% relative improvement). Besides, NIKI obtains comparable performance to pixel-aligned approaches, showing a 1.2 mm improvement on PA-MPJPE.

Tab.~\ref{table:3dpw_xocc} demonstrates the robustness of NIKI to extreme occlusions and truncations. We report the results of the most accurate pixel-aligned and direct regression approaches on the \texttt{3DPW-XOCC} dataset. It shows that direct regression approaches outperform pixel-aligned approaches in challenging scenes, which is in contrast to the results in the standard benchmark. NIKI improves the PA-MPJPE performance by \textbf{38.7}\% compared to HybrIK and \textbf{10.1}\% compared to PARE that finetuned on the \texttt{3DPW-XOCC} train set.

\begin{table}[t]
    \begin{center}
        \resizebox{\linewidth}{!}
        {
            \begin{tabular}{l|cccc}

            \toprule
			 & \multicolumn{4}{c}{ \texttt{AGORA} }  \\
			\cmidrule(lr){2-5}
            Method & ~NMVE~$\downarrow$~ & ~NMJE~$\downarrow$~ & ~MVE~$\downarrow$~ & ~MPJPE~$\downarrow$~ \\
            \midrule
            HMR~\cite{hmr} & 217.0 & 226.0 & 173.6 & 180.5 \\
            SPIN~\cite{spin} & 216.3 & 223.1 & 168.7 & 175.1 \\
            EFT~\cite{joo2021exemplar} & 196.3 & 203.6 & 159.0 & 165.4 \\
            ROMP~\cite{sun2021monocular} & 227.3 & 236.6 & 161.4 & 168.0 \\
            PyMAF~\cite{zhang2021pymaf} & 200.2 & 207.4 & 168.2 & 174.2 \\
            PARE~\cite{kocabas2021pare} & 167.7 & 174.0 & 140.9 & 146.2 \\
            SPEC~\cite{kocabas2021spec} & 126.8 & 133.7 & 106.5 & 112.3 \\
            CLIFF~\cite{li2022cliff} & 83.5 & 89.0 & 76.0 & 81.0 \\
            HybrIK~\cite{li2021hybrik} & 81.2 & 84.6 & 73.9 & 77.0 \\
            \midrule
            NIKI (One-Stage) & 72.2 & 75.9 & 65.7 & 69.1 \\
            NIKI (Twist-and-Swing) & \textbf{70.2} & \textbf{74.0} & \textbf{63.9} & \textbf{67.3} \\

            \bottomrule
            \end{tabular}
        }
        \caption{\textbf{Quantitative comparisons with state-of-the-art methods on the \texttt{AGORA} dataset.}}
        \label{table:agora}
    \end{center}
\end{table}

\begin{table}[t]
	\centering
	\resizebox{\columnwidth}{!}
    {
		\begin{tabular}{l|c|c|c|c|c}
			\toprule
			& \multicolumn{3}{c|}{ \texttt{3DPW-OCC} } & \multicolumn{2}{c}{ \texttt{3DOH} } \\
			\cmidrule(lr){2-6}
			{Method} & {\footnotesize MPJPE $\downarrow$} & {\footnotesize PA-MPJPE $\downarrow$} & {\footnotesize PVE $\downarrow$} & {\footnotesize MPJPE $\downarrow$} & {\footnotesize PA-MPJPE $\downarrow$}  \\
			\midrule
			Zhang~\etal~\cite{zhang2020object} & - & 72.2 & - & - & 58.5 \\
			SPIN~\cite{spin} & 98.4 & 62.5 & 135.1 & 104.3 & 68.3 \\
			HMR-EFT~\cite{joo2021exemplar} & 95.8 & 62.0 & 120.5 & 75.2 & 53.1 \\
			HybrIK~\cite{li2021hybrik} & 90.8 & 58.8 & 111.9 & 40.4 & 31.2 \\
			PARE~\cite{kocabas2021pare} & 91.4 & 57.4 & 115.3 & 63.3 & 44.3 \\
			\midrule
            NIKI (One-Stage) & 88.2 & 55.3 & 109.7 & 38.9 & 29.2 \\
            NIKI (Twist-and-Swing) & \textbf{85.5} & \textbf{53.5} & \textbf{107.6} & \textbf{38.8} & \textbf{28.7} \\
			\bottomrule
		\end{tabular}
	}
	\caption{\textbf{Quantitative comparisons with state-of-the-art methods on the \texttt{3DPW-OCC} and \texttt{3DOH} datasets.}}
	\label{table:etc_occ}
\end{table}{}

The results of NIKI on other occlusion-specific datasets are reported in Tab.~\ref{table:agora} and Tab.~\ref{table:etc_occ}. NIKI shows consistent improvements on all these datasets, demonstrating that NIKI is robust to challenging occlusions and truncations while maintaining pixel-aligned accuracy. Specifically, NIKI improves the NMJE performance on \texttt{AGORA} by \textbf{12.5}\% compared to the state-of-the-art methods. We can also observe that the twist-and-swing mapping model is consistently superior to the one-stage mapping model. More discussions of limitations and future work are provided in the supplementary material.

% Although NIKI shows superior performance across all benchmarks, it has two limitations. First, NIKI does not include body shape refinement. The pose positions contain the bone length information and can help refine $\boldsymbol{\beta}$ for better body shape estimation. Second, NIKI does not use the scene information to separate the pose error. Using scene constraints can reduce implausible human-scene interactions and further improve robustness. A detailed discussion is provided in the supplementary material. We believe these limitations are exciting avenues for future work to explore.

\subsection{Ablation Study}

\paragraph{INN \vs  NN.}
To demonstrate the superiority of the invertible network in solving the IK problem, we compare NIKI with existing IK algorithms, including the analytical~\cite{li2021hybrik}, MLP-based~\cite{zhou2019continuity}, the INN-based \cite{ardizzone2018analyzing}, and the vanilla INN baseline. \cite{ardizzone2018analyzing} is trained without modeling the error information. The vanilla INN baseline is trained in both the forward and inverse directions without independence and boundary constraints. Quantitative results are reported in Tab.~\ref{table:ablation}a. The MLP-based method shows better performance in occlusions scenarios compared to the analytical method. However, it is less accurate in the standard non-occlusion scenarios since it cannot accurately track the movement of the body joints. The vanilla INN performs better than the MLP model in both standard and occlusion datasets but is still less accurate than the analytical method in the standard dataset. NIKI surpasses all methods in both datasets.

\begin{figure}[t]
    \centering
    \includegraphics[width=0.96\linewidth]{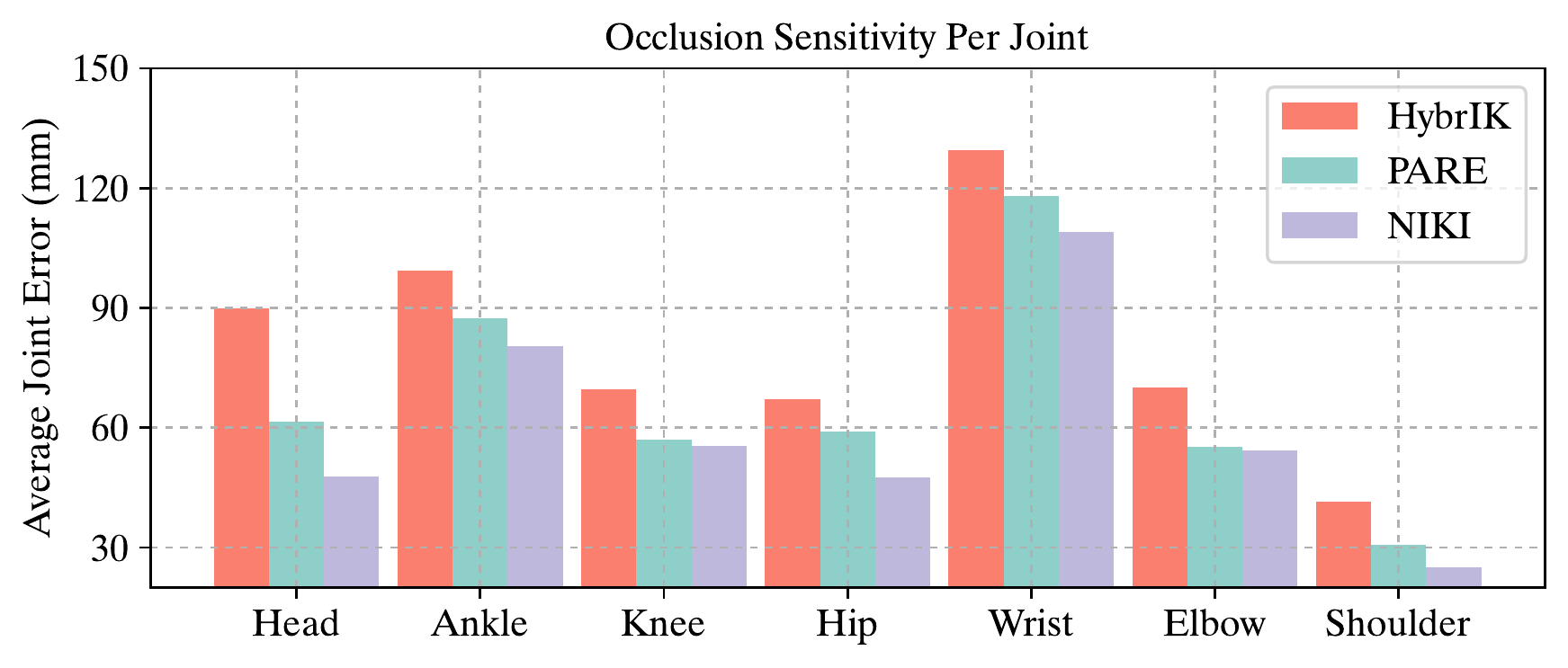}
 
    \caption{\textbf{Per joint occlusion sensitivity analysis} of three different methods: HybrIK~\cite{li2021hybrik}, PARE~\cite{kocabas2021pare}, and NIKI.}
    \label{fig:sensitive}
\end{figure}

\begin{table}[t]
	\centering
	\resizebox{\columnwidth}{!}
    {
		\begin{tabular}{ll|c|c|c|c}
			\toprule
			& & \multicolumn{2}{c|}{ \texttt{3DPW} } & \multicolumn{2}{c}{ \texttt{3DPW-XOCC} } \\
			\cmidrule(lr){3-6}
			& & {\footnotesize MPJPE $\downarrow$} & {\footnotesize PA-MPJPE $\downarrow$} & {\footnotesize MPJPE $\downarrow$} & {\footnotesize PA-MPJPE $\downarrow$} \\
            \midrule
			{\multirow{3}{*}{{(a)}}} & Analytical IK~\cite{li2021hybrik} & 74.9 & 41.8 & 148.3 & 98.7 \\
            & MLP-based IK~\cite{zhou2019continuity} & 83.2 & 50.3 & 121.1 & 68.5  \\
            & Vanilla INN-based IK & 73.8 & 43.1 & 115.3 & 64.4 \\
			& Ardizzone \etal~\cite{ardizzone2018analyzing} & 79.1 & 45.6 & 119.5 & 67.3 \\
            \midrule 
            {\multirow{4}{*}{{(b)}}} & NIKI & \textbf{71.3} & \textbf{40.6} & \textbf{110.7} & \textbf{60.5} \\
			& w/o Independence & 71.6 & 40.8 & 112.6 & 61.4 \\
            & w/o Boundary & 73.6 & 43.0 & 112.5 & 62.9 \\
			% \midrule
            & w/o Bi-directional Train & 74.0 & 43.0 & 111.9 & 61.9 \\
			\bottomrule
		\end{tabular}
	}
	\caption{\textbf{Ablation experiments on the \texttt{3DPW} and \texttt{3DPW-XOCC} datasets.}}
	\label{table:ablation}
\end{table}

\paragraph{Effectiveness of Error Decoupling.}
To study the effectiveness of error decoupling, we evaluate models that are trained without enforcing independence or boundary constraints.
% The error embedding $\epsilon_r$ could contains pose information without the independence constraint.
% We also report the results of the model without boundary constraints.
Quantitative results are summarized in Tab.~\ref{table:ablation}b. It shows that the independence loss for inverse error modeling contributes to a better performance in occlusion scenarios and barely affects the performance in non-occlusion scenarios. Besides, the boundary constraints contribute to a better alignment. Without boundary constraints, the model shows performance degradation in non-occlusion scenarios.

\begin{figure*}[t]
    \centering
    \includegraphics[width=0.96\linewidth]{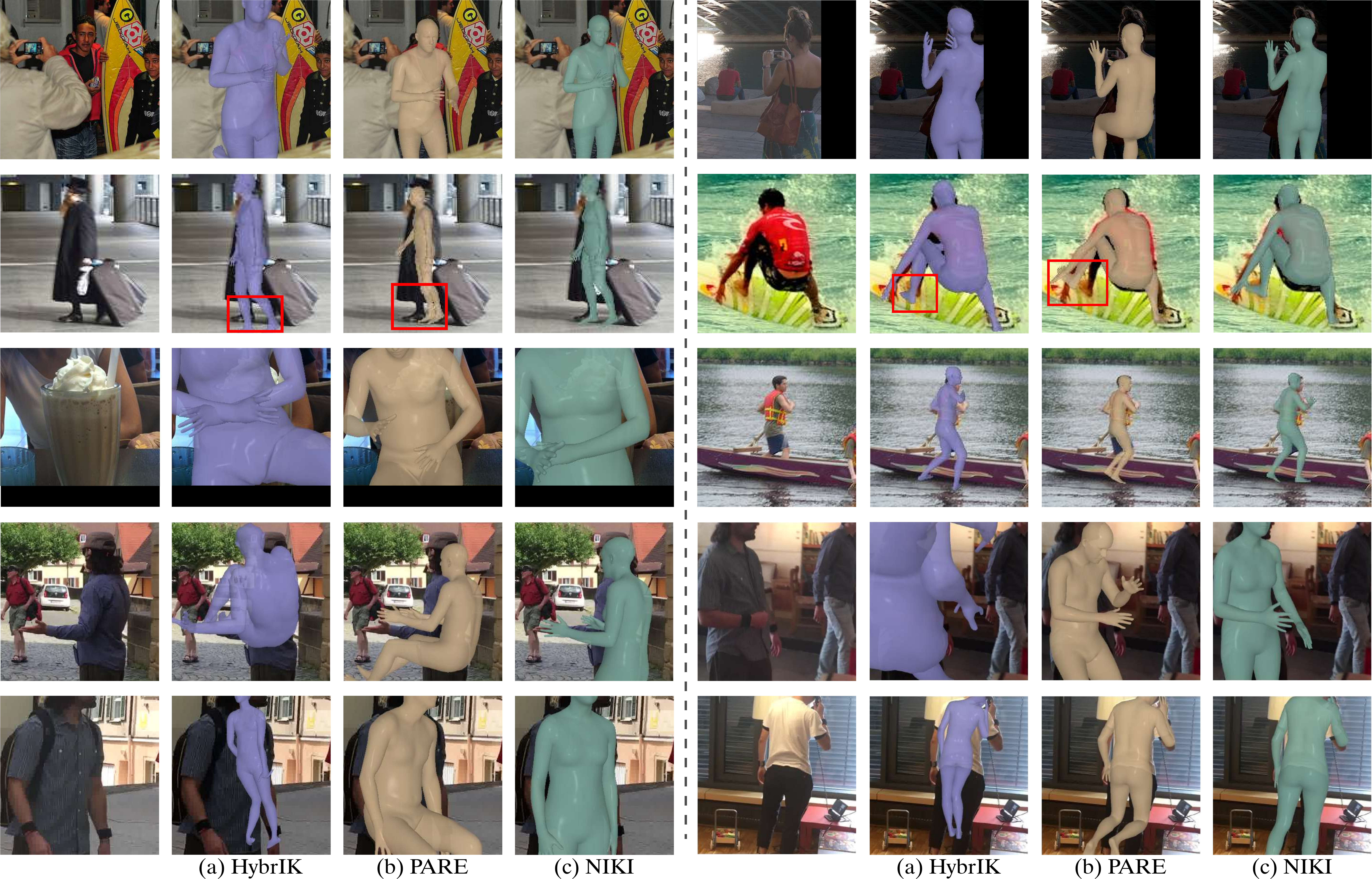}
 
    \caption{\textbf{Quantitative results on COCO (rows 1-3) and 3DPW-XOCC (rows 4-5) datasets.} From left to right: Input image, (a) HybrIK~\cite{li2021hybrik} results, (b) PARE~\cite{kocabas2021pare} results, and (c) NIKI results.}
    \label{fig:qualitative}
\end{figure*}

\paragraph{Effectiveness of Bi-directional Training.}
To further validate the effectiveness of bi-directional training, we report the results of the baseline model that is only trained with the inverse process. Without bi-directional training, we also cannot apply the boundary condition in the forward direction, which means that we only decouple the errors in the inverse process. As shown in Tab.~\ref{table:ablation}b, the IK model cannot maintain the sensitivity to non-occluded body joints in the standard benchmark without forward training.

\paragraph{Sensitivity Analysis.}
We further follow Kocabas \etal~\cite{kocabas2021pare} to conduct the occlusion sensitivity analysis. Fig.~\ref{fig:sensitive} shows the per-joint breakdown of the mean 3D error from the occlusion sensitivity analysis for three different methods on the \texttt{3DPW} test split. Although HybrIK~\cite{li2021hybrik} obtains high accuracy on the \texttt{3DPW} dataset, it is quite sensitive to occlusions. NIKI is more robust to occlusions and improves the robustness of all joints. We also qualitatively compare HybrIK, PARE, and NIKI in Fig.~\ref{fig:qualitative}. NIKI performs well in challenging occlusion scenarios and predicts well-aligned results. More occlusion analyses and qualitative samples are provided in the supplementary material.

% \subsection{Limitations and Future Work.}
% Although NIKI shows superior performance across all benchmarks, it has two limitations. First, NIKI does not include body shape refinement. The pose positions contain the bone length information and can help refine $\boldsymbol{\beta}$ for better body shape estimation. Second, NIKI does not use the scene information to separate the pose error. Using scene constraints can reduce implausible human-scene interactions and further improve robustness. A detailed discussion is provided in the supplementary material. We believe these limitations are exciting avenues for future work to explore.

% \paragraph{Qualitative Comparison.}
% We qualitatively compare HybrIK, PARE and NIKI in Fig.~\ref{fig:qualitative}.

\section{Conclusion}
In this paper, we propose NIKI, a neural inverse kinematics solution for accurate and robust 3D human pose and shape estimation. NIKI is built with invertible neural networks to model bi-directional error information in the forward and inverse kinematics processes.
% To leverage the pixel-aligned local evidences and keep robust to occlusions, we introduce bi-directional error modeling using the characteristics of the bijective and invertible mappings of INN. 
% NIKI 
% We leverage the characteristics of the bijective and invertible mappings of INN to improve the model robustness to occlusions while maintaining the pixel-aligned accuracy in daily non-occlusion scenes.
In the inverse direction, NIKI explicitly decouples the error information from the manifold of the plausible human poses to improve robustness. In the forward direction, NIKI enforces zero-error boundaries to obtain accurate mesh-image alignment. We construct the invertible neural network by emulating the analytical inverse kinematics algorithm with twist-and-swing decomposition to improve interpretability.
% We improve the interpretability of the invertible network by analogizing the analytical IK algorithm with twist-and-swing decompisition.
Comprehensive experiments on standard and occlusion-specific datasets demonstrate the pixel-aligned accuracy and robustness of NIKI. We hope NIKI can serve as a solid baseline for challenging real-world applications.

% \paragraph{Limitations} Our method has two limitations: i) not including scene information; ii) not including body shape refinement. A detailed discussion is provided in the supplementary material. We believe these limitations are exciting avenues for future work to explore.

\noindent\textbf{Acknowledgments.} This work was supported by the National Key R\&D Program of China (No. 2021ZD0110704), Shanghai Municipal Science and Technology Major Project (2021SHZDZX0102), Shanghai Qi Zhi Institute, and Shanghai Science and Technology Commission (21511101200).

\appendix

\section*{Appendix}\label{sec:appendix}

\section{Architecture of INN}
\label{sec:arch}

\begin{figure}[h]
    \centering
    \includegraphics[width=\linewidth]{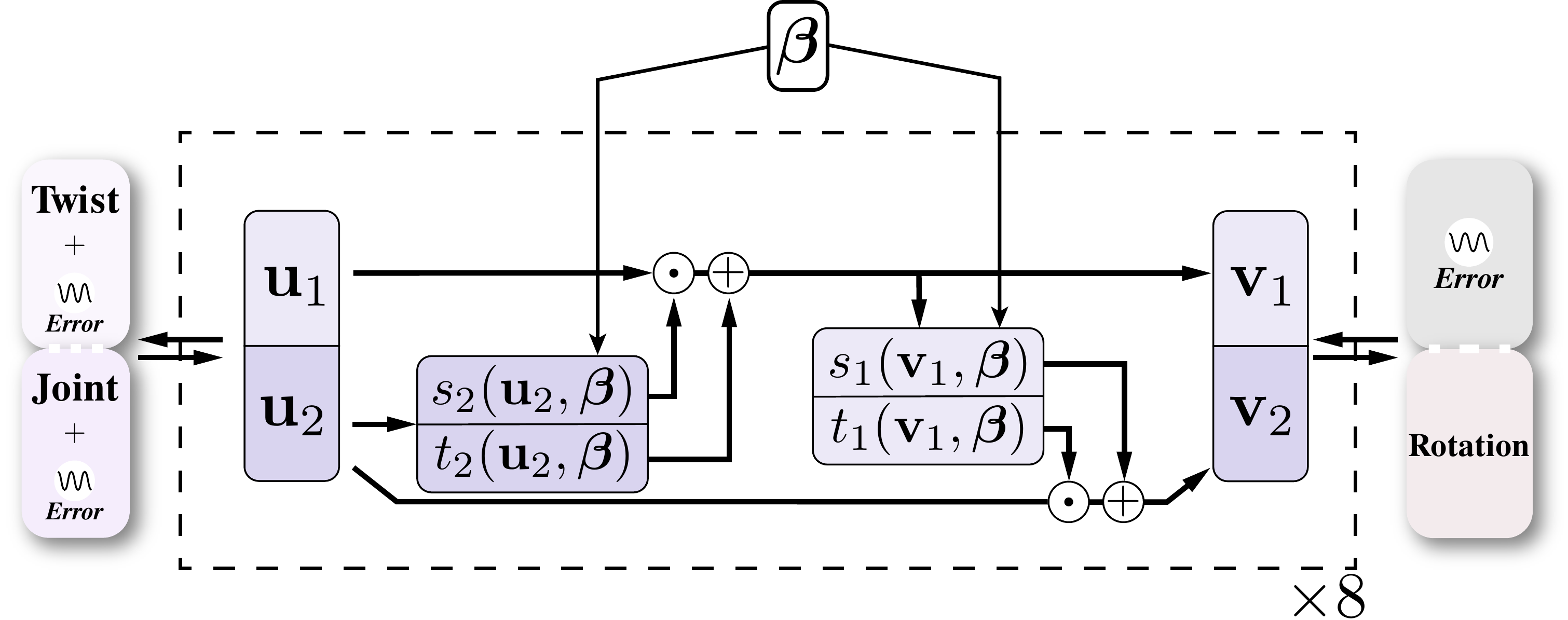}

    \caption{\textbf{Detailed architecture of the one-stage mapping model.}}
    \label{fig:arch_one}
 \end{figure}

\subsection{One-Stage Mapping}

The detailed architecture of the one-stage mapping model is illustrated in Fig.~\ref{fig:arch_one}. We follow the architecture of RealNVP~\cite{dinh2016density}. The model consists of multiple basic blocks to increase capacity. The input vector $\mathbf{u}$ of the block is split into two parts, $\mathbf{u}_1$ and $\mathbf{u}_2$, which are subsequently transformed with coefficients $\exp(s_i)$ and $t_i$ ($i \in \{1, 2\}$) by the two affine coupling layers:
\begin{align}
    \mathbf{v}_1 = \mathbf{u}_1 \odot \exp(s_2(\mathbf{u}_2, \boldsymbol{\beta})) + t_2(\mathbf{u}_2, \boldsymbol{\beta}), \label{eq:affine1}\\
    \mathbf{v}_2 = \mathbf{u}_2 \odot \exp(s_1(\mathbf{v}_1, \boldsymbol{\beta})) + t_1(\mathbf{v}_1, \boldsymbol{\beta}), \label{eq:affine2}
\end{align}
where $\mathbf{v} = [\mathbf{v}_1, \mathbf{v}_2]$ is the output vector of the block and $\odot$ denotes element-wise multiplication. The coefficients of the affine transformation can be learned by arbitrarily complex functions, which do not need to be invertible. The invertibility is guaranteed by the affine transformation in Eq.~\ref{eq:affine1} and \ref{eq:affine2}. The scale network $s_i$ is a 3-layer MLP with the hidden dimension of $512$, and the translation network $t_i$ has the same architecture followed by a $\tanh$ activation function.

\begin{figure}[t]
    \centering
    \includegraphics[width=\linewidth]{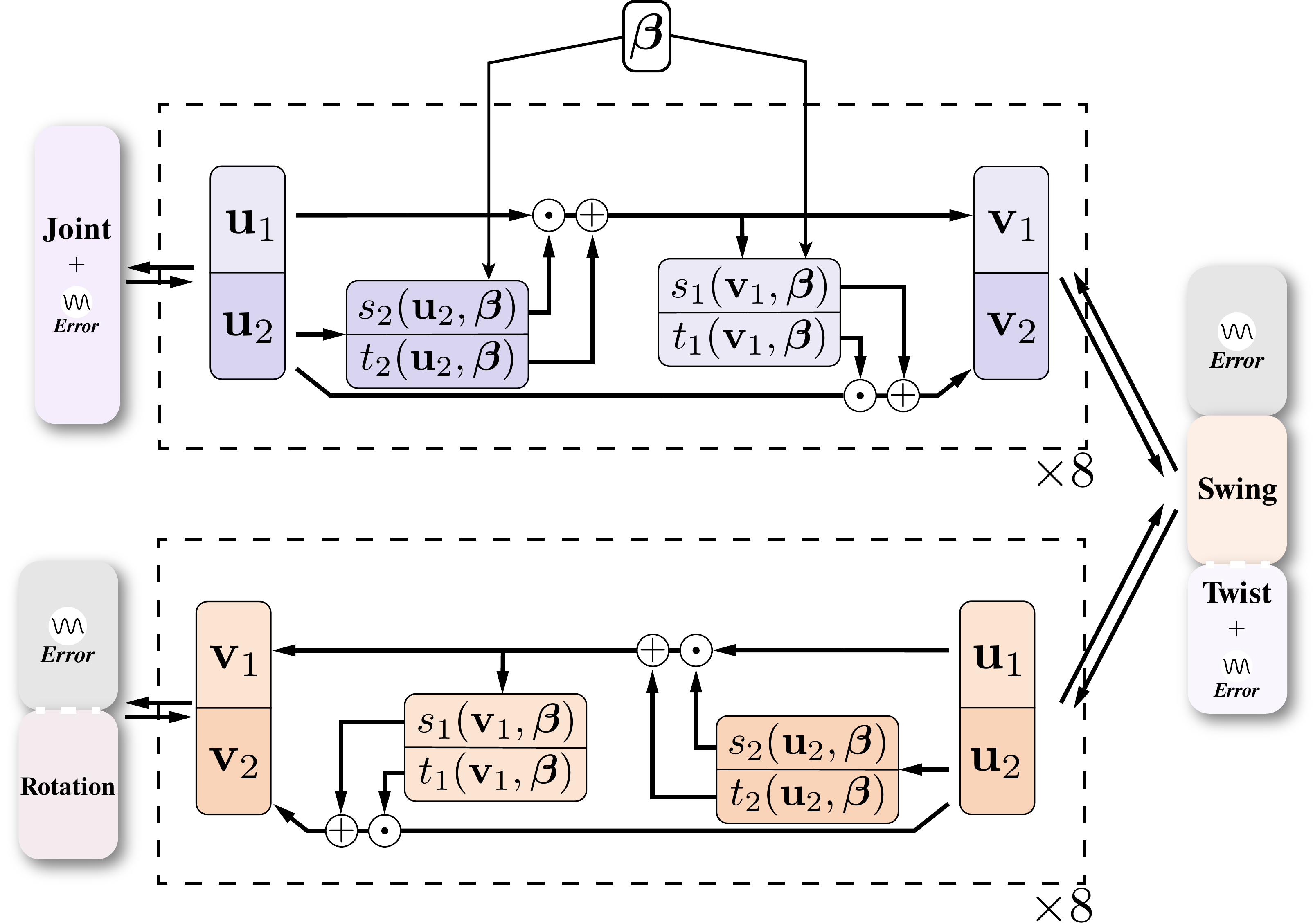}

    \caption{\textbf{Detailed architecture of the twist-and-swing mapping model.}}
    \label{fig:arch_two}
 \end{figure}

\subsection{Twist-and-Swing Mapping}
The detailed architecture of the twist-and-swing mapping model is illustrated in Fig.~\ref{fig:arch_two}. The two-step mapping is implemented by two separate invertible networks. The first network has the same architecture as the one-stage mapping model, while its input is only the joint positions, and the output is the swing rotations. The second network removes the shape condition and directly transforms the twist and swing rotations to complete rotations.

\section{Implementation Details}
\label{sec:supp_impl}

In our experiments, we use the weights pretrained on \texttt{COCO}~\cite{coco} 2D pose estimation task for the initialization of the CNN backbone to accelerate convergence. The scalar coefficients in the loss function are $\lambda_{\textit{inv}}=1$, $\lambda_{\textit{fwd}}=1$, $\lambda_{\textit{ind}}=1$, $\lambda_{\textit{bnd}}^i=0.1$, $\lambda_{\textit{bnd}}^f=1$. We first train the CNN backbone following HybrIK~\cite{li2021hybrik} to obtain initial joint positions and twist rotations. Then we solely train NIKI and freeze the parameters of the CNN backbone. During training, we follow EFT~\cite{joo2021exemplar}, SPIN~\cite{spin}, and PARE~\cite{kocabas2021pare}, which use fixed data sampling ratios for each batch. We incorporate 50\% \texttt{Human3.6M} and 50\% \texttt{3DPW} when conducting experiments on the \texttt{3DPW} and \texttt{3DPW-XOCC} datasets. For experiments on the \texttt{3DPW-OCC} and \texttt{3DOH} datasets, we incorporate 35\% \texttt{COCO}, 35\% \texttt{Human3.6M}, and 30\% \texttt{3DOH}.

\begin{figure*}[t]
    \centering
    \includegraphics[width=0.9\linewidth]{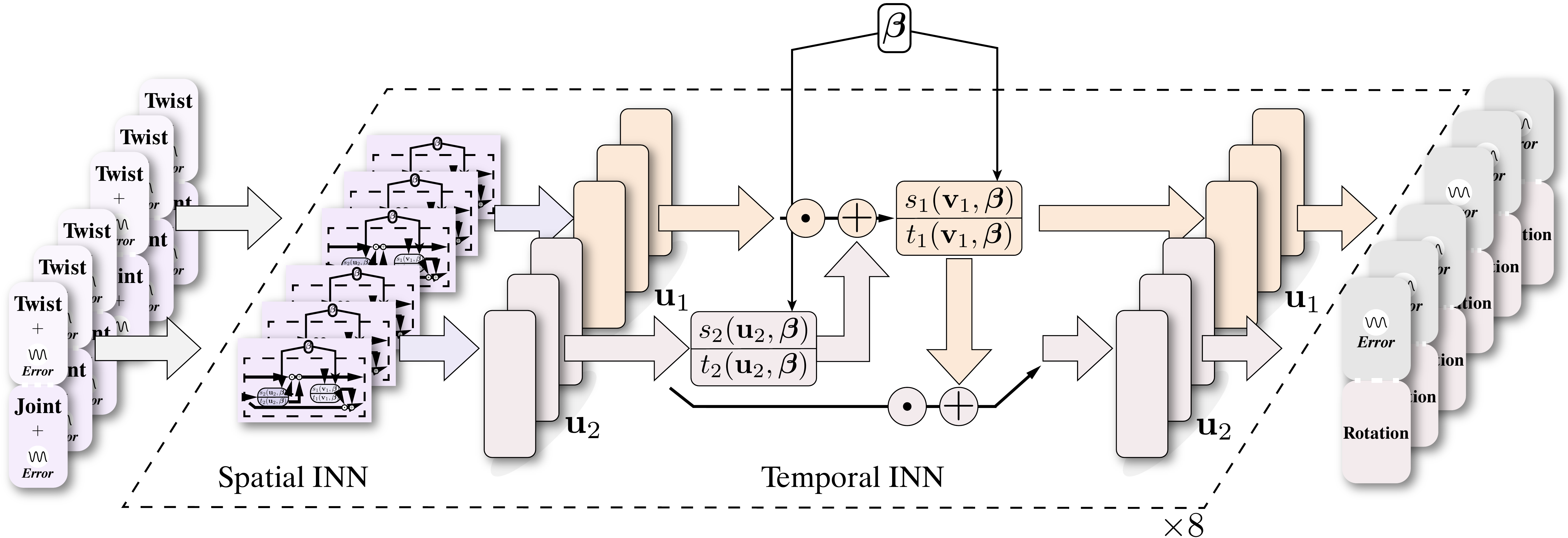}

    \caption{\textbf{Detailed architecture of the temporal INN.}}
    \label{fig:time}
\end{figure*}

\section{Temporal Extension of NIKI}
\label{sec:time}

\subsection{Architecture}

We extend the invertible network for temporal input. We design a spatial-temporal INN model to incorporate temporal information to solve the IK problem. For simplicity, we use the basic block in the one-stage mapping and twist-and-swing mapping models as the spatial INN. Self-attention modules are introduced to serve as the temporal INN and conduct temporal affine transformations. The temporal input vectors $\{\mathbf{u}^t\}_{1}^{T}$ are split into two subsets, $\{\mathbf{u}^{t}\}_{1}^{\lfloor T/2 \rfloor}$ and $\{\mathbf{u}^{t}\}_{\lfloor T/2 \rfloor + 1}^{T}$, which are subsequently transformed with coefficients $\exp(s_i)$ and $t_i$ ($i \in \{1, 2\}$) by the two affine coupling layers like Eq.~\ref{eq:affine1} and \ref{eq:affine2}. We adopt self-attention layers~\cite{vaswani2017attention} as the temporal scale and translation layers. The detailed network architecture of the temporal INN is illustrated in Fig.~\ref{fig:time}.

\begin{table}[t]
    \begin{center}
        \resizebox{\linewidth}{!}
        {
            \begin{tabular}{l|cccc}

            \toprule
			& \multicolumn{4}{c}{ \texttt{3DPW} }  \\
			\cmidrule(lr){2-5}
            Method & ~MPJPE~$\downarrow$~ & ~PA-MPJPE~$\downarrow$~ & ~PVE~$\downarrow$~ & ~ACCEL~$\downarrow$ \\
            \midrule
            % \parbox[t]{2mm}{\multirow{5}{*}{\rotatebox[origin=c]{90}{Temporal}}} &
            % HMMR~\cite{hmr} & 116.5 & 72.6 & - \\
            VIBE~\cite{vibe} & 82.9 & 51.9 & 99.1 & 23.4 \\
            MEVA~\cite{luo20203d} & 86.9 & 54.7 & - & 11.6 \\
            TCMR~\cite{choi2021beyond} & 86.5 & 52.7 & 102.9 & 7.1 \\
            MAED~\cite{wan2021encoder} & 79.1 & 45.7 & 92.6 & 17.6 \\
            D\&D~\cite{li2022d} & 73.7 & 42.7 & 88.6 & \textbf{7.0} \\
            % \midrule
            % \parbox[t]{2mm}{\multirow{5}{*}{\rotatebox[origin=c]{90}{Regression}}} & HMR~\cite{hmr} & 130.0 & 81.3 & - \\
            % & SPIN~\cite{spin} & 96.9 & 59.2 & 116.4 \\
            % & ROMP~\cite{sun2021monocular} & 85.5 & 53.3 & 103.1 \\
            % & METRO~\cite{lin2021end} & 77.1 & 47.9 & 88.2 \\
            % & PARE~\cite{kocabas2021pare} & 74.5 & 46.5 & 88.6 \\
            % \midrule
            % \parbox[t]{2mm}{\multirow{6}{*}{\rotatebox[origin=c]{90}{Pixel-aligned}}} & PyMAF~\cite{zhang2021pymaf} & 92.8 & 58.9 & 110.1  \\
            % & I2L~\cite{moon2020i2l} & 93.2 & 58.6 & - \\
            % & KAMA~\cite{iqbal2021kama} & - & 51.1 & 97.0 \\
            % & Mesh Graphormer~\cite{lin2021mesh} & 74.7 & 45.6 & 87.7 \\
            % & HybrIK (ResNet)~\cite{li2021hybrik} & 76.2 & 45.1 & 89.1 \\
            % & HybrIK (HRNet)~\cite{li2021hybrik} & 72.9 & 41.8 & 88.6 \\
			% % \cmidrule(lr){2-5}
            \midrule
            NIKI (Frame-based) & {71.3} & {40.6} & {86.6} & 15.1 \\
            NIKI (Temporal) & \textbf{71.2} & \textbf{40.5} & \textbf{86.3} & 12.3 \\
            \bottomrule
            \end{tabular}
        }
        \caption{\textbf{Quantitative comparisons with state-of-the-art temporal methods on the \texttt{3DPW} dataset.} Symbol ``-'' means results are not available.}
        \label{table:time_3dpw}
    \end{center}
\end{table}

\begin{table}[t]
    \begin{center}
        \resizebox{\linewidth}{!}
        {
            \begin{tabular}{l|cccc}

            \toprule
			 & \multicolumn{4}{c}{ \texttt{3DPW-XOCC} }  \\
			\cmidrule(lr){2-5}
            Method & ~MPJPE~$\downarrow$~ & ~PA-MPJPE~$\downarrow$~ & ~PVE~$\downarrow$~ & ~ACCEL~$\downarrow$ \\
            \midrule
            HybrIK~\cite{li2021hybrik} & 148.3 & 98.7 & 164.5 & 108.6 \\
            % HybrIK~\cite{li2021hybrik} + Transformer~\cite{vaswani2017attention} & 148.3 & 98.7 & 164.5 \\
            PARE$^*$~\cite{kocabas2021pare} & 114.2 & 67.7 & 133.0 & 90.7 \\
            PARE$^*$~\cite{kocabas2021pare} + VIBE~\cite{vibe} & 97.3 & 60.2 & 114.9 & 18.3 \\
            \midrule
            % NIKI (One-Stage) & {117.0} & {64.6} & 135.6 \\
            NIKI (Frame-based) & 110.7 & 60.5 & 128.6 & 74.4 \\
            NIKI (Temporal) & \textbf{88.9} & \textbf{52.1} & \textbf{98.0} & \textbf{17.3} \\

            \bottomrule
            \end{tabular}
        }
        \caption{\textbf{Quantitative comparisons with state-of-the-art temporal methods on the \texttt{3DPW-XOCC} dataset.} Symbol $*$ means finetuning on the \texttt{3DPW-XOCC} train set.}
        \label{table:time_3dpw_xocc}
    \end{center}
\end{table}

\begin{figure}[t]
    \centering
    % \fbox{\rule{0pt}{1in} \rule{0.9\linewidth}{0pt}}
    \includegraphics[width=\linewidth]{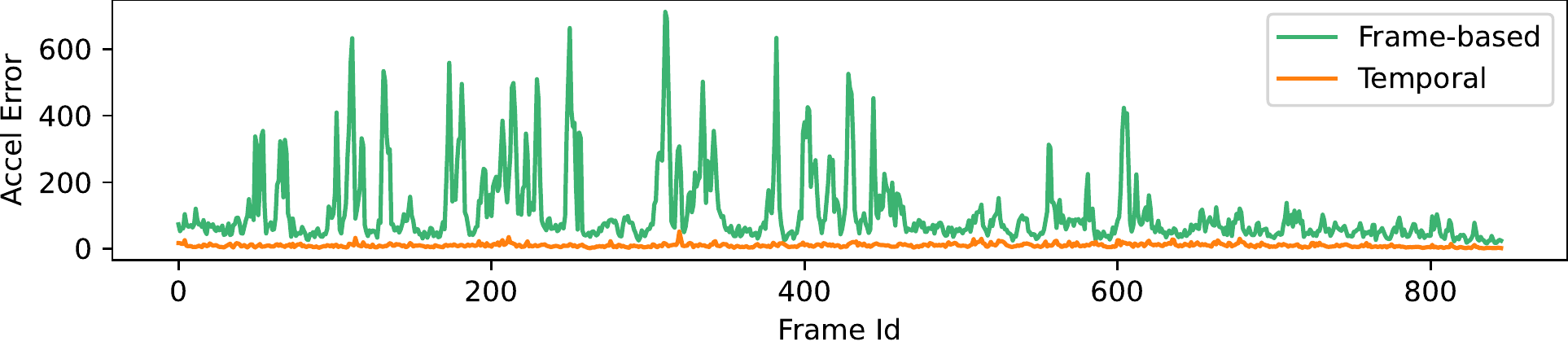}
    \caption{\textbf{Acceleration error curve.}}
    \label{fig:accl}
\end{figure}

% \section{Experiments}
% \label{sec:supp_exp}

\subsection{Experiments of the Temporal Extension}

We evaluate the temporal extension on both standard and occlusion-specific benchmarks. Tab.~\ref{table:time_3dpw} compares temporal NIKI with previous state-of-the-art temporal HPS methods on the standard \texttt{3DPW}~\cite{3dpw} dataset. Notice that we do not design complex network architecture or use dynamics information. Our temporal extension simply applies the affine coupling layers to the time domain. It shows that our simple extension obtains better accuracy than state-of-the-art dynamics-based approaches.

Tab.~\ref{table:time_3dpw_xocc} presents the performance on the occlusion-specific benchmark. We compare the temporal extension with a strong baseline. The baseline combines PARE~\cite{kocabas2021pare} with the state-of-the-art temporal approach, VIBE~\cite{vibe}. We first use the backbone of PARE~\cite{kocabas2021pare} to extract attention-guided features. Then we apply VIBE~\cite{vibe} to incorporate temporal information to predict smooth and robust human motions. Temporal NIKI outperforms the baseline in challenging occlusions and truncations.

Fig.~\ref{fig:accl} present the acceleration error curves of the single-frame and temporal models in the \texttt{3DPW-XOCC} dataset. We can observe that the temporal model can improve motion smoothness.

\begin{figure}[t]
    \centering
    % \fbox{\rule{0pt}{8in} \rule{0.9\linewidth}{0pt}}
    \includegraphics[width=0.9\linewidth]{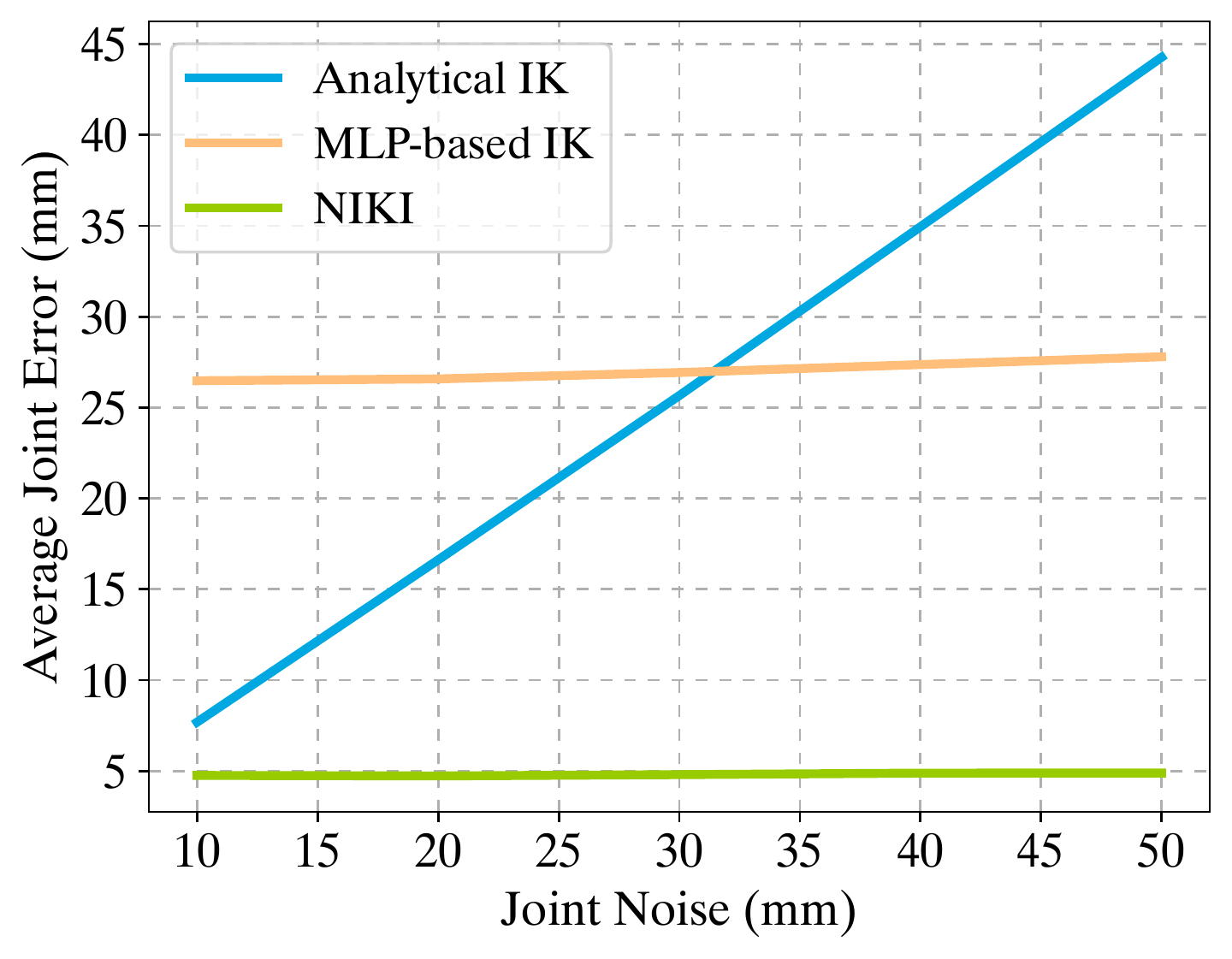}

    \caption{\textbf{Noise sensitivity analysis} of analytical IK, MLP-based IK and NIKI.}
    \label{fig:noise}
\end{figure}

\section{Noise Analysis}
\label{sec:noise}

We assess the robustness of three different IK algorithms: analytical IK, MLP-based IK, and NIKI. We evaluate their performance on the \texttt{AMASS} dataset~\cite{mahmood2019amass} with noisy joint positions. As shown in Fig.~\ref{fig:noise}, MLP-based IK is more robust than the analytical IK when the noise is larger than 30 mm. However, MLP-based IK fails to obtain pixel-aligned performance when the noise is small. NIKI shows superior performance at all noise levels.

\begin{figure}[!t]
    \centering
    % \fbox{\rule{0pt}{1in} \rule{0.9\linewidth}{0pt}}
    \includegraphics[width=\linewidth]{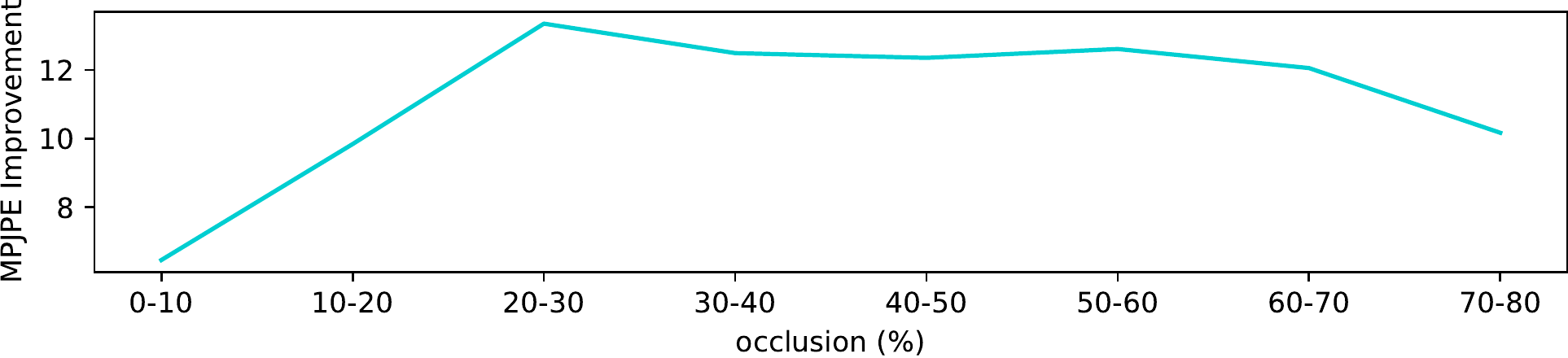}
    \caption{\textbf{Improvement over different occlusion levels.}}
    \label{fig:occ_agora}
 \end{figure}

\section{Collision Analysis}
\label{sec:collision}

To quantitatively show that the output poses from NIKI are more plausible, we compare the collision ratio of mesh triangles~\cite{mihajlovic2022coap} between HybrIK and NIKI on the \texttt{3DPW-XOCC} dataset.  NIKI reduces the collision ratio
from 2.6\% to 1.0\% (57.7\% relative improvement).

\section{Occlusion Analysis}
\label{sec:occlusion}

\begin{figure*}[t]
    \centering
    % \fbox{\rule{0pt}{8in} \rule{0.9\linewidth}{0pt}}
    \includegraphics[width=\linewidth]{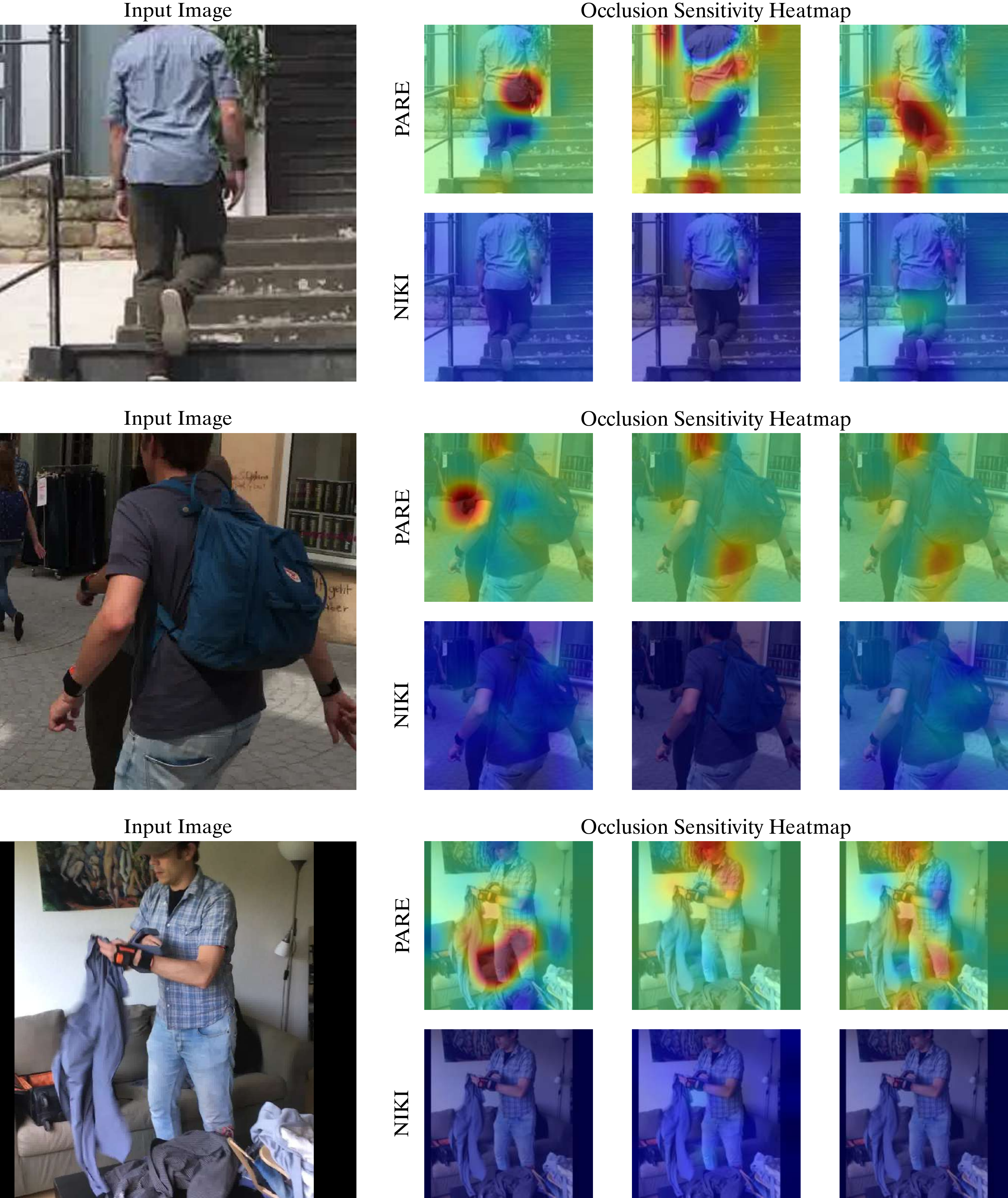}

    \caption{\textbf{Occlusion Sensitivity Maps of PARE~\cite{kocabas2021pare} and NIKI.}}
    \label{fig:occ_n_p}
\end{figure*}

\begin{figure*}[t]
    \centering
    % \fbox{\rule{0pt}{8in} \rule{0.9\linewidth}{0pt}}
    \includegraphics[width=\linewidth]{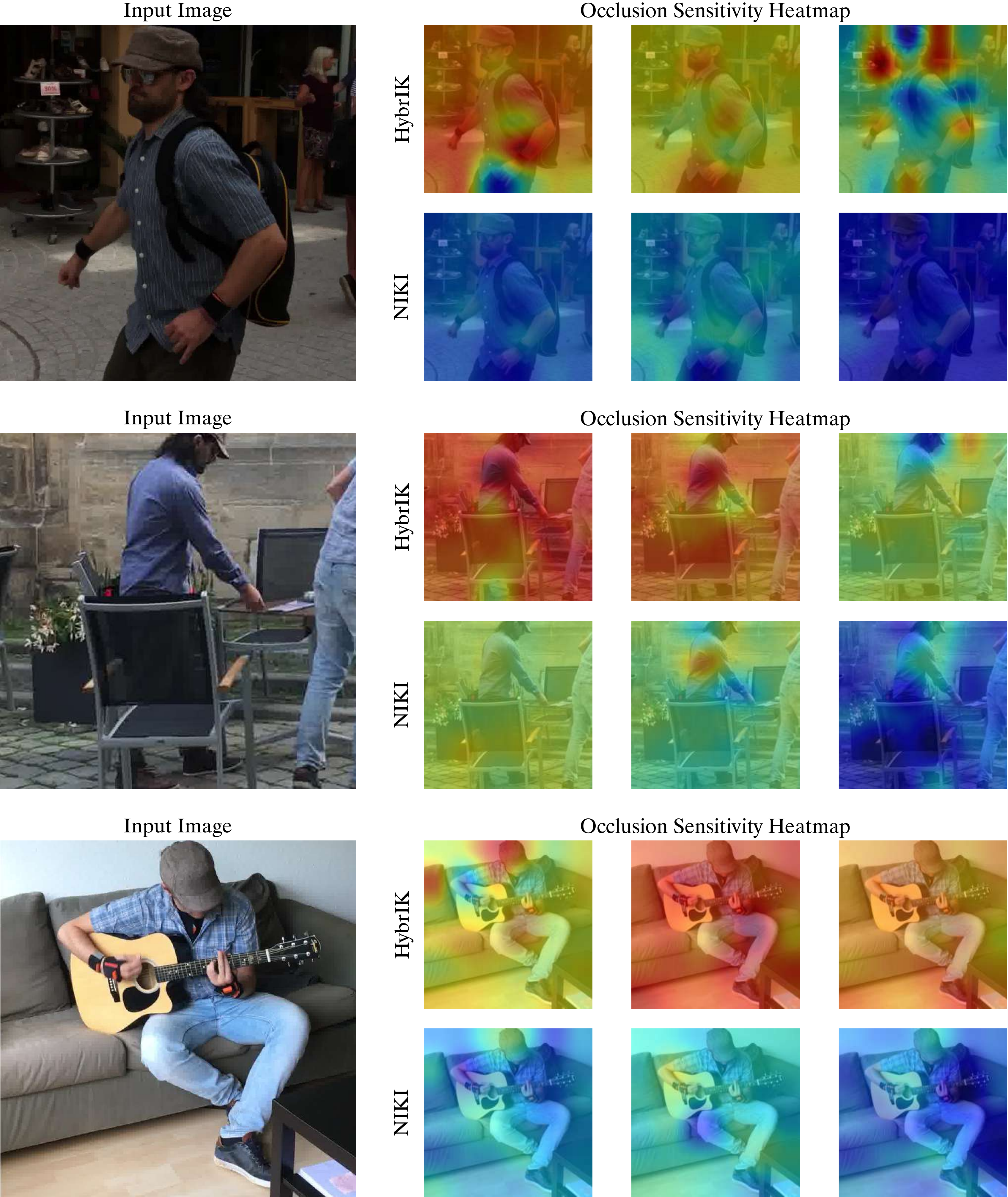}

    \caption{\textbf{Occlusion Sensitivity Maps of HybrIK~\cite{li2021hybrik} and NIKI.}}
    \label{fig:occ_n_h}
\end{figure*}

We follow the framework of \cite{zeiler2014visualizing,kocabas2021pare} and replace the classification score with an error measure for body poses. We choose MPJPE as the error measurement. This analysis is not limited to a particular network architecture. We apply it to the state-of-the-art pixel-aligned approach, HybrIK~\cite{li2021hybrik}, and the direct regression approach, PARE~\cite{kocabas2021pare}. The visualizations of the error maps are shown in Fig.~\ref{fig:occ_n_p} and \ref{fig:occ_n_h}. Warmer colors denote a higher MPJPE. It shows that NIKI is more robust to body part occlusions.

Additionally, we follow the official AGORA analyses to compare the performance in different occlusion levels. As shown in Fig.~\ref{fig:occ_agora}, in the low occlusion level (0-10\%), NIKI brings $6.5$ mm MPJPE improvement. The improvement reaches a peak ($13.3$ mm) in the medium occlusion level (20-30\%). For the high occlusion level (70-80\%), the improvement falls back to $10.2$ mm. We can observe that NIKI is good at handling medium occlusions. There is still a lot of room for improvement in highly occluded scenarios.

\begin{table}[t]
	\centering
	\resizebox{\columnwidth}{!}
    {
		\begin{tabular}{l|c|c|c|c}
			\toprule
			& \multicolumn{2}{c|}{ \texttt{3DPW} } & \multicolumn{2}{c}{ \texttt{3DPW-XOCC} } \\
			\cmidrule(lr){2-5}
			& {\footnotesize MPJPE $\downarrow$} & {\footnotesize PA-MPJPE $\downarrow$} & {\footnotesize MPJPE $\downarrow$} & {\footnotesize PA-MPJPE $\downarrow$} \\
            \midrule
            NIKI & 71.3 & 40.6 & \textbf{110.7} & \textbf{60.5} \\
			$+$ Heatmap Cond.~\cite{wehrbein2021probabilistic} & \textbf{71.1} & \textbf{40.4} & 110.8 & 60.6 \\
			\bottomrule
		\end{tabular}
	}
	\caption{\textbf{Integrate heatmap condition.}}
	\label{table:cond}
\end{table}

\section{Heatmap Condition}
\label{sec:cond}

We follow Wehrbein \etal~\cite{wehrbein2021probabilistic} and add heatmap condition in the INN. As shown in Tab.~\ref{table:cond}, it brings $0.2$ mm improvement on the \texttt{3DPW} dataset. However, it is $0.1$ mm worse on the \texttt{3DPW-XOCC} dataset. We assume this is because heatmap is not reliable under server occlusions.

\section{Inference Time and Model Size}
\label{sec:complexity}

We benchmark the inference time of the analytical IK algorithm, HybrIK~\cite{li2021hybrik} and NIKI with an RTX 3090 GPU with a batch size of 1. The latency of HybrIK is $26$ ms and NIKI is $8$ ms, respectively. HybrIK is much slower since it needs to solve the rotations iteratively along the kinematic tree. For the model size, the total parameters of NIKI is 29.01M.

\section{Details of \texttt{3DPW-XOCC}}
\label{sec:xocc}

\texttt{3DPW-XOCC} is a new benchmark for human pose and shape estimation with extremely challenging occlusions and truncations. The dataset is augmented from the original \texttt{3DPW} dataset by adding temporally-smooth synthetic occlusions and truncations. To ensure temporal smoothness, we choose keyframes at an interval of 8 frames, and the rest frames are generated by linearly interpolating the clipping and occlusion of the keyframes. In the keyframe, the image is randomly clipped to ensure that at least one body part is outside the clipped image with a possibility of over $2/3$. A square area that takes up to 30\% of the clipped image is replaced by gaussian noise to serve as occlusion. The evaluation protocol and the split of the dataset are unchanged.

\section{Limitations and Future Work}
\label{sec:limitation}

Our work has several limitations. First, NIKI does not include body shape refinement. Human body shape estimation is also challenging in occlusion scenarios. The incorrect body shape would cause incorrect distal joints reconstruction. For example, even the knee and ankle rotations are correct, the wrong leg length will cause a wrong ankle position. Exploiting the bone length information in joint positions can help refine $\boldsymbol{\beta}$ for better pose and shape estimation.
% The pose positions contain the bone length information. Therefore, using the bone length can help refine $\boldsymbol{\beta}$ for better body shape estimation.
Second, NIKI does not use the scene information to separate the pose error. The initial joint positions could be physiologically plausible but do not match the input scene. Using scene constraints can reduce implausible human-scene interactions and further improve robustness. Third, the training of NIKI relies on the diversity of datasets. To accurately built the bijective mapping, the training data need to be diverse enough. We believe these limitations are exciting avenues for future work to explore.

\begin{figure*}[t]
    \centering
    % \fbox{\rule{0pt}{8in} \rule{0.9\linewidth}{0pt}}
    \includegraphics[width=\linewidth]{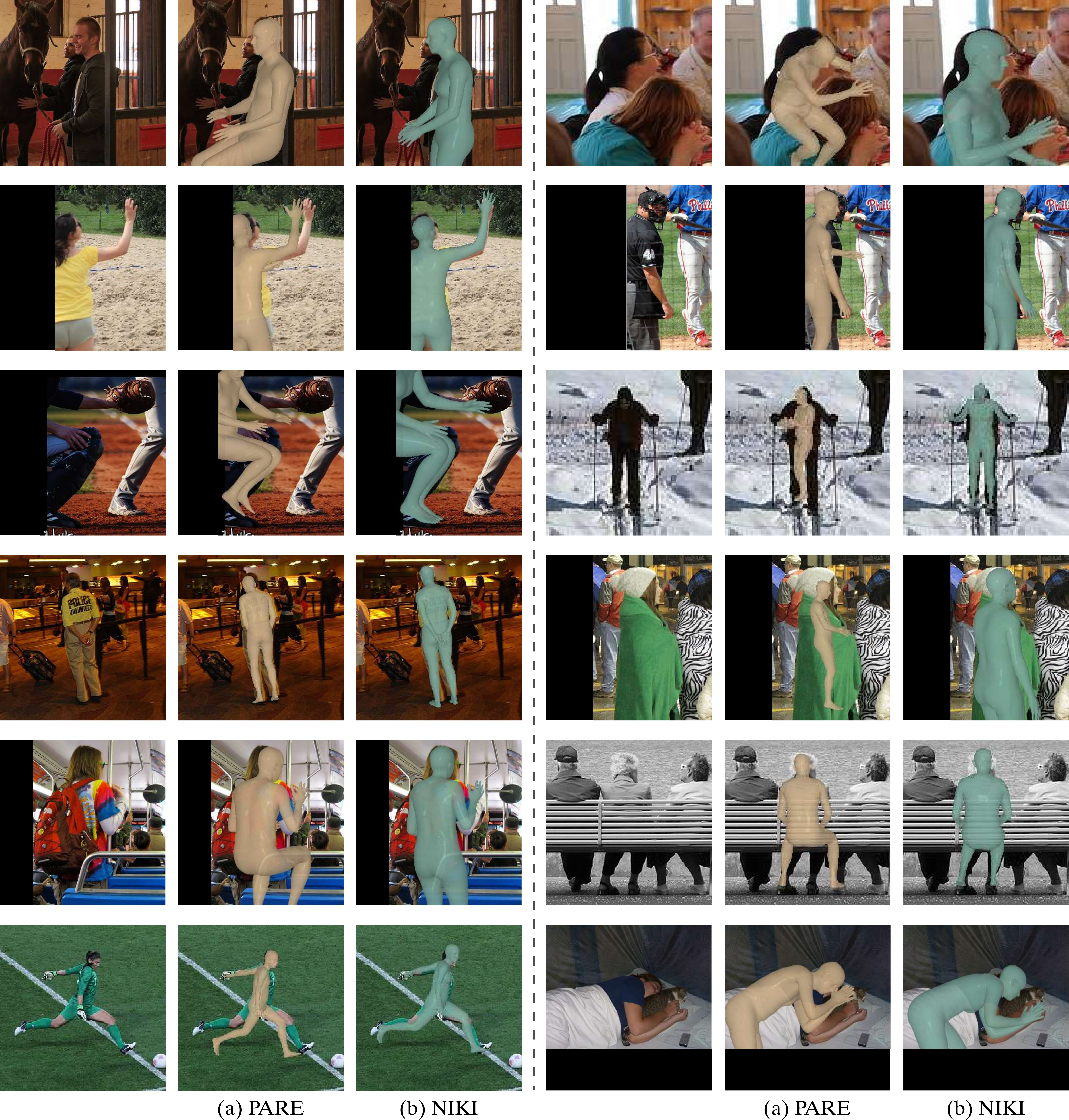}

    \caption{\textbf{Qualitative comparison with PARE~\cite{kocabas2021pare}.}}
    \label{fig:qualitative_w_pare}
\end{figure*}

\begin{figure*}[t]
    \centering
    % \fbox{\rule{0pt}{8in} \rule{0.9\linewidth}{0pt}}
    \includegraphics[width=\linewidth]{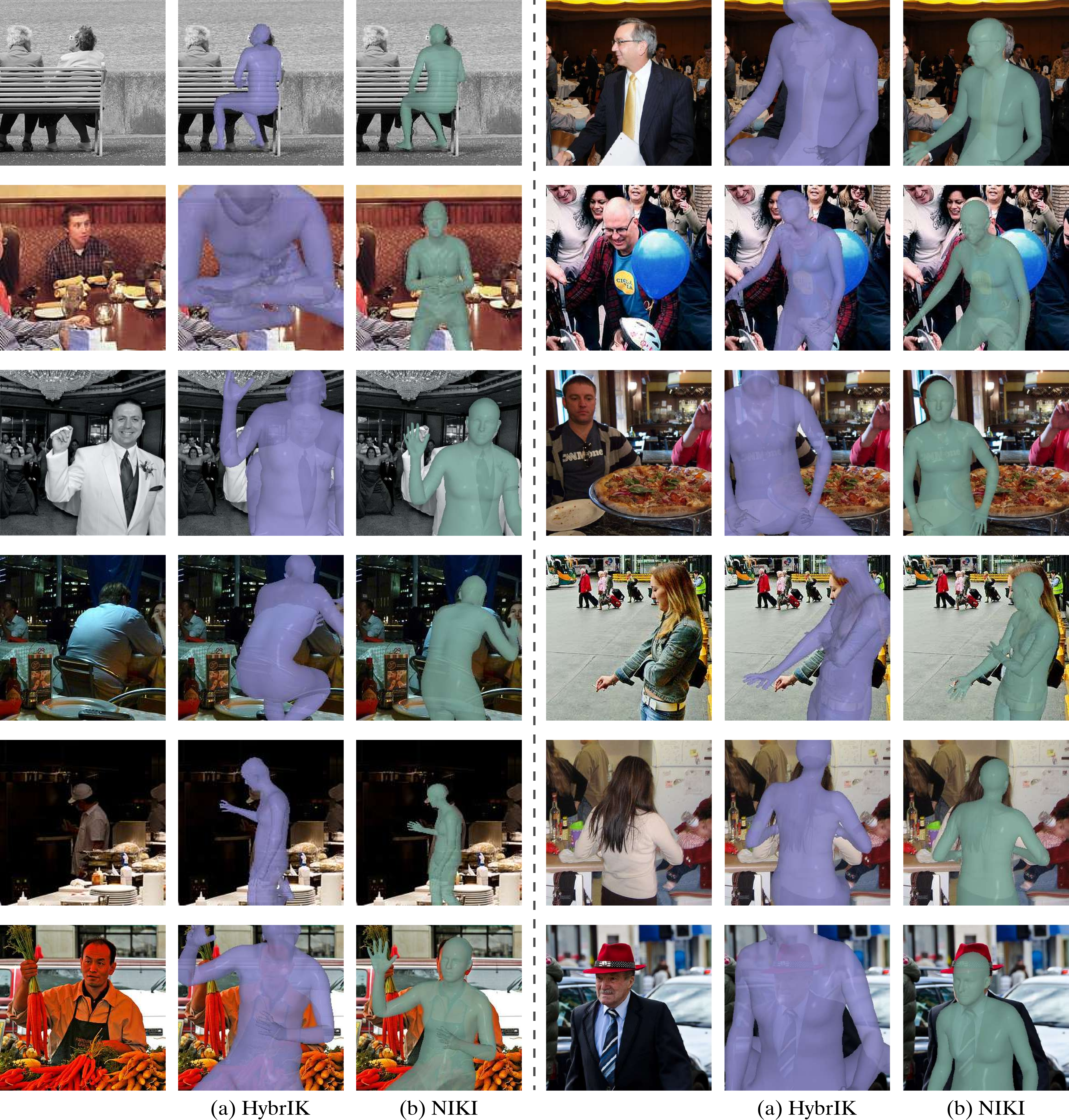}

    \caption{\textbf{Qualitative comparison with HybrIK~\cite{li2021hybrik}.}}
    \label{fig:qualitative_w_hybrik}
\end{figure*}

\section{Qualitative Results}
\label{sec:qualitative}

Additional qualitative results are shown in Fig.~\ref{fig:qualitative_w_pare} and \ref{fig:qualitative_w_hybrik}.

\newpage
\twocolumn[
\begin{center}
\addcontentsline{toc}{section}{References}
\end{center}
]
%%%%%%%%% REFERENCES
{\small
\bibliographystyle{ieee_fullname}
\bibliography{egbib}

\begin{thebibliography}{10}\itemsep=-1pt

\bibitem{anguelov2005scape}
Dragomir Anguelov, Praveen Srinivasan, Daphne Koller, Sebastian Thrun, Jim
  Rodgers, and James Davis.
\newblock Scape: shape completion and animation of people.
\newblock In {\em SIGGRAPH}, 2005.

\bibitem{ardizzone2018analyzing}
Lynton Ardizzone, Jakob Kruse, Sebastian Wirkert, Daniel Rahner, Eric~W
  Pellegrini, Ralf~S Klessen, Lena Maier-Hein, Carsten Rother, and Ullrich
  K{\"o}the.
\newblock Analyzing inverse problems with invertible neural networks.
\newblock In {\em ICLR}, 2019.

\bibitem{aristidou2011fabrik}
Andreas Aristidou and Joan Lasenby.
\newblock Fabrik: A fast, iterative solver for the inverse kinematics problem.
\newblock {\em Graphical Models}, 2011.

\bibitem{balestrino1984robust}
A Balestrino, Giuseppe De~Maria, and L Sciavicco.
\newblock Robust control of robotic manipulators.
\newblock {\em IFAC Proceedings Volumes}, 1984.

\bibitem{biggs20203d}
Benjamin Biggs, David Novotny, Sebastien Ehrhardt, Hanbyul Joo, Ben Graham, and
  Andrea Vedaldi.
\newblock 3d multi-bodies: Fitting sets of plausible 3d human models to
  ambiguous image data.
\newblock In {\em NeurIPS}, 2020.

\bibitem{bogo2016keep}
Federica Bogo, Angjoo Kanazawa, Christoph Lassner, Peter Gehler, Javier Romero,
  and Michael~J Black.
\newblock Keep it smpl: Automatic estimation of 3d human pose and shape from a
  single image.
\newblock In {\em ECCV}, 2016.

\bibitem{buss2005selectively}
Samuel~R Buss and Jin-Su Kim.
\newblock Selectively damped least squares for inverse kinematics.
\newblock {\em Journal of Graphics tools}, 2005.

\bibitem{choi2021beyond}
Hongsuk Choi, Gyeongsik Moon, Ju~Yong Chang, and Kyoung~Mu Lee.
\newblock Beyond static features for temporally consistent 3d human pose and
  shape from a video.
\newblock In {\em CVPR}, 2021.

\bibitem{csiszar2017solving}
Akos Csiszar, Jan Eilers, and Alexander Verl.
\newblock On solving the inverse kinematics problem using neural networks.
\newblock In {\em M2VIP}, 2017.

\bibitem{dai2023sloper4d}
Yudi Dai, YiTai Lin, XiPing Lin, Chenglu Wen, Lan Xu, Hongwei Yi, Siqi Shen,
  Yuexin Ma, and Cheng Wang.
\newblock Sloper4d: A scene-aware dataset for global 4d human pose estimation
  in urban environments.
\newblock In {\em CVPR}, 2023.

\bibitem{dinh2016density}
Laurent Dinh, Jascha Sohl-Dickstein, and Samy Bengio.
\newblock Density estimation using real nvp.
\newblock In {\em ICLR}, 2017.

\bibitem{girard1985computational}
Michael Girard and Anthony~A Maciejewski.
\newblock Computational modeling for the computer animation of legged figures.
\newblock In {\em SIGGRAPH}, 1985.

\bibitem{gretton2012kernel}
Arthur Gretton, Karsten~M Borgwardt, Malte~J Rasch, Bernhard Sch{\"o}lkopf, and
  Alexander Smola.
\newblock A kernel two-sample test.
\newblock {\em JMLR}, 2012.

\bibitem{h36m}
Catalin Ionescu, Dragos Papava, Vlad Olaru, and Cristian Sminchisescu.
\newblock Human3. 6m: Large scale datasets and predictive methods for 3d human
  sensing in natural environments.
\newblock {\em TPAMI}, 2013.

\bibitem{iqbal2021kama}
Umar Iqbal, Kevin Xie, Yunrong Guo, Jan Kautz, and Pavlo Molchanov.
\newblock Kama: 3d keypoint aware body mesh articulation.
\newblock In {\em 3DV}, 2021.

\bibitem{jiang2020coherent}
Wen Jiang, Nikos Kolotouros, Georgios Pavlakos, Xiaowei Zhou, and Kostas
  Daniilidis.
\newblock Coherent reconstruction of multiple humans from a single image.
\newblock In {\em CVPR}, 2020.

\bibitem{joo2021exemplar}
Hanbyul Joo, Natalia Neverova, and Andrea Vedaldi.
\newblock Exemplar fine-tuning for 3d human model fitting towards in-the-wild
  3d human pose estimation.
\newblock In {\em 3DV}, 2021.

\bibitem{hmr}
Angjoo Kanazawa, Michael~J Black, David~W Jacobs, and Jitendra Malik.
\newblock End-to-end recovery of human shape and pose.
\newblock In {\em CVPR}, 2018.

\bibitem{kanazawa2019learning}
Angjoo Kanazawa, Jason~Y Zhang, Panna Felsen, and Jitendra Malik.
\newblock Learning 3d human dynamics from video.
\newblock In {\em CVPR}, 2019.

\bibitem{khirodkar2022occluded}
Rawal Khirodkar, Shashank Tripathi, and Kris Kitani.
\newblock Occluded human mesh recovery.
\newblock In {\em CVPR}, 2022.

\bibitem{klein1983review}
Charles~A Klein and Ching-Hsiang Huang.
\newblock Review of pseudoinverse control for use with kinematically redundant
  manipulators.
\newblock {\em IEEE Transactions on Systems, Man, and Cybernetics}, 1983.

\bibitem{vibe}
Muhammed Kocabas, Nikos Athanasiou, and Michael~J Black.
\newblock Vibe: Video inference for human body pose and shape estimation.
\newblock In {\em CVPR}, 2020.

\bibitem{kocabas2021pare}
Muhammed Kocabas, Chun-Hao~P Huang, Otmar Hilliges, and Michael~J Black.
\newblock Pare: Part attention regressor for 3d human body estimation.
\newblock In {\em ICCV}, 2021.

\bibitem{kocabas2021spec}
Muhammed Kocabas, Chun-Hao~P Huang, Joachim Tesch, Lea M{\"u}ller, Otmar
  Hilliges, and Michael~J Black.
\newblock Spec: Seeing people in the wild with an estimated camera.
\newblock In {\em ICCV}, 2021.

\bibitem{spin}
Nikos Kolotouros, Georgios Pavlakos, Michael~J Black, and Kostas Daniilidis.
\newblock Learning to reconstruct 3d human pose and shape via model-fitting in
  the loop.
\newblock In {\em ICCV}, 2019.

\bibitem{kolotouros2021probabilistic}
Nikos Kolotouros, Georgios Pavlakos, Dinesh Jayaraman, and Kostas Daniilidis.
\newblock Probabilistic modeling for human mesh recovery.
\newblock In {\em ICCV}, 2021.

\bibitem{lassner2017unite}
Christoph Lassner, Javier Romero, Martin Kiefel, Federica Bogo, Michael~J
  Black, and Peter~V Gehler.
\newblock Unite the people: Closing the loop between 3d and 2d human
  representations.
\newblock In {\em CVPR}, 2017.

\bibitem{li2022d}
Jiefeng Li, Siyuan Bian, Chao Xu, Gang Liu, Gang Yu, and Cewu Lu.
\newblock D\&d: Learning human dynamics from dynamic camera.
\newblock In {\em ECCV}, 2022.

\bibitem{li2021human}
Jiefeng Li, Siyuan Bian, Ailing Zeng, Can Wang, Bo Pang, Wentao Liu, and Cewu
  Lu.
\newblock Human pose regression with residual log-likelihood estimation.
\newblock In {\em ICCV}, 2021.

\bibitem{li2021hybrik}
Jiefeng Li, Chao Xu, Zhicun Chen, Siyuan Bian, Lixin Yang, and Cewu Lu.
\newblock Hybrik: A hybrid analytical-neural inverse kinematics solution for 3d
  human pose and shape estimation.
\newblock In {\em CVPR}, 2021.

\bibitem{li2022cliff}
Zhihao Li, Jianzhuang Liu, Zhensong Zhang, Songcen Xu, and Youliang Yan.
\newblock Cliff: Carrying location information in full frames into human pose
  and shape estimation.
\newblock In {\em ECCV}, 2022.

\bibitem{liao2023car}
Tingting Liao, Xiaomei Zhang, Yuliang Xiu, Hongwei Yi, Xudong Liu, Guo-Jun Qi,
  Yong Zhang, Xuan Wang, Xiangyu Zhu, and Zhen Lei.
\newblock {High-Fidelity Clothed Avatar Reconstruction from a Single Image}.
\newblock In {\em CVPR}, 2023.

\bibitem{lin2021end}
Kevin Lin, Lijuan Wang, and Zicheng Liu.
\newblock End-to-end human pose and mesh reconstruction with transformers.
\newblock In {\em CVPR}, 2021.

\bibitem{lin2021mesh}
Kevin Lin, Lijuan Wang, and Zicheng Liu.
\newblock Mesh graphormer.
\newblock In {\em ICCV}, 2021.

\bibitem{coco}
Tsung-Yi Lin, Michael Maire, Serge Belongie, James Hays, Pietro Perona, Deva
  Ramanan, Piotr Doll{\'a}r, and C~Lawrence Zitnick.
\newblock Microsoft coco: Common objects in context.
\newblock In {\em ECCV}, 2014.

\bibitem{loper2015smpl}
Matthew Loper, Naureen Mahmood, Javier Romero, Gerard Pons-Moll, and Michael~J
  Black.
\newblock Smpl: A skinned multi-person linear model.
\newblock {\em TOG}, 2015.

\bibitem{luenberger1984linear}
David~G Luenberger, Yinyu Ye, et~al.
\newblock {\em Linear and nonlinear programming}.
\newblock Springer, 1984.

\bibitem{luo20203d}
Zhengyi Luo, S~Alireza Golestaneh, and Kris~M Kitani.
\newblock 3d human motion estimation via motion compression and refinement.
\newblock In {\em ACCV}, 2020.

\bibitem{mahmood2019amass}
Naureen Mahmood, Nima Ghorbani, Nikolaus~F Troje, Gerard Pons-Moll, and
  Michael~J Black.
\newblock Amass: Archive of motion capture as surface shapes.
\newblock In {\em ICCV}, 2019.

\bibitem{mihajlovic2022coap}
Marko Mihajlovic, Shunsuke Saito, Aayush Bansal, Michael Zollhoefer, and Siyu
  Tang.
\newblock Coap: Compositional articulated occupancy of people.
\newblock In {\em CVPR}, 2022.

\bibitem{moon2020i2l}
Gyeongsik Moon and Kyoung~Mu Lee.
\newblock I2l-meshnet: Image-to-lixel prediction network for accurate 3d human
  pose and mesh estimation from a single rgb image.
\newblock In {\em ECCV}, 2020.

\bibitem{oreshkin2021protores}
Boris~N Oreshkin, Florent Bocquelet, Felix~G Harvey, Bay Raitt, and Dominic
  Laflamme.
\newblock Protores: Proto-residual network for pose authoring via learned
  inverse kinematics.
\newblock In {\em ICLR}, 2021.

\bibitem{osmansupr}
Ahmed~AA Osman, Timo Bolkart, Dimitrios Tzionas, and Michael~J Black.
\newblock Supr: A sparse unified part-based human representation.
\newblock In {\em ECCV}, 2022.

\bibitem{patel2021agora}
Priyanka Patel, Chun-Hao~P. Huang, Joachim Tesch, David~T. Hoffmann, Shashank
  Tripathi, and Michael~J. Black.
\newblock {AGORA}: Avatars in geography optimized for regression analysis.
\newblock In {\em CVPR}, 2021.

\bibitem{pavlakos2019expressive}
Georgios Pavlakos, Vasileios Choutas, Nima Ghorbani, Timo Bolkart, Ahmed~AA
  Osman, Dimitrios Tzionas, and Michael~J Black.
\newblock Expressive body capture: 3d hands, face, and body from a single
  image.
\newblock In {\em CVPR}, 2019.

\bibitem{pavlakos2018learning}
Georgios Pavlakos, Luyang Zhu, Xiaowei Zhou, and Kostas Daniilidis.
\newblock Learning to estimate 3d human pose and shape from a single color
  image.
\newblock In {\em CVPR}, 2018.

\bibitem{peng2021neural}
Sida Peng, Yuanqing Zhang, Yinghao Xu, Qianqian Wang, Qing Shuai, Hujun Bao,
  and Xiaowei Zhou.
\newblock Neural body: Implicit neural representations with structured latent
  codes for novel view synthesis of dynamic humans.
\newblock In {\em CVPR}, 2021.

\bibitem{rempe2021humor}
Davis Rempe, Tolga Birdal, Aaron Hertzmann, Jimei Yang, Srinath Sridhar, and
  Leonidas~J Guibas.
\newblock Humor: 3d human motion model for robust pose estimation.
\newblock In {\em ICCV}, 2021.

\bibitem{rokbani2015ik}
Nizar Rokbani, Alicia Casals, and Adel~M Alimi.
\newblock Ik-fa, a new heuristic inverse kinematics solver using firefly
  algorithm.
\newblock In {\em Computational intelligence applications in modeling and
  control}. Springer, 2015.

\bibitem{sun2021monocular}
Yu Sun, Qian Bao, Wu Liu, Yili Fu, Michael~J Black, and Tao Mei.
\newblock Monocular, one-stage, regression of multiple 3d people.
\newblock In {\em ICCV}, 2021.

\bibitem{tiwari2022pose}
Garvita Tiwari, Dimitrije Anti{\'c}, Jan~Eric Lenssen, Nikolaos Sarafianos,
  Tony Tung, and Gerard Pons-Moll.
\newblock Pose-ndf: Modeling human pose manifolds with neural distance fields.
\newblock In {\em ECCV}, 2022.

\bibitem{varol2018bodynet}
Gul Varol, Duygu Ceylan, Bryan Russell, Jimei Yang, Ersin Yumer, Ivan Laptev,
  and Cordelia Schmid.
\newblock Bodynet: Volumetric inference of 3d human body shapes.
\newblock In {\em ECCV}, 2018.

\bibitem{vaswani2017attention}
Ashish Vaswani, Noam Shazeer, Niki Parmar, Jakob Uszkoreit, Llion Jones,
  Aidan~N Gomez, {\L}ukasz Kaiser, and Illia Polosukhin.
\newblock Attention is all you need.
\newblock {\em NeurIPS}, 2017.

\bibitem{villegas2018neural}
Ruben Villegas, Jimei Yang, Duygu Ceylan, and Honglak Lee.
\newblock Neural kinematic networks for unsupervised motion retargetting.
\newblock In {\em CVPR}, 2018.

\bibitem{voleti2022smpl}
Vikram Voleti, Boris~N Oreshkin, Florent Bocquelet, F{\'e}lix~G Harvey,
  Louis-Simon M{\'e}nard, and Christopher Pal.
\newblock Smpl-ik: Learned morphology-aware inverse kinematics for ai driven
  artistic workflows.
\newblock In {\em SIGGRAPH Asia}, 2022.

\bibitem{3dpw}
Timo von Marcard, Roberto Henschel, Michael~J Black, Bodo Rosenhahn, and Gerard
  Pons-Moll.
\newblock Recovering accurate 3d human pose in the wild using imus and a moving
  camera.
\newblock In {\em ECCV}, 2018.

\bibitem{wampler1986manipulator}
Charles~W Wampler.
\newblock Manipulator inverse kinematic solutions based on vector formulations
  and damped least-squares methods.
\newblock {\em IEEE Transactions on Systems, Man, and Cybernetics}, 1986.

\bibitem{wan2021encoder}
Ziniu Wan, Zhengjia Li, Maoqing Tian, Jianbo Liu, Shuai Yi, and Hongsheng Li.
\newblock Encoder-decoder with multi-level attention for 3d human shape and
  pose estimation.
\newblock In {\em ICCV}, 2021.

\bibitem{wehrbein2021probabilistic}
Tom Wehrbein, Marco Rudolph, Bodo Rosenhahn, and Bastian Wandt.
\newblock Probabilistic monocular 3d human pose estimation with normalizing
  flows.
\newblock In {\em ICCV}, 2021.

\bibitem{wolovich1984computational}
William~A Wolovich and H Elliott.
\newblock A computational technique for inverse kinematics.
\newblock In {\em CDC}, 1984.

\bibitem{xiu2022econ}
Yuliang Xiu, Jinlong Yang, Xu Cao, Dimitrios Tzionas, and Michael~J Black.
\newblock Econ: Explicit clothed humans obtained from normals.
\newblock In {\em CVPR}, 2023.

\bibitem{xiu2022icon}
Yuliang Xiu, Jinlong Yang, Dimitrios Tzionas, and Michael~J Black.
\newblock Icon: implicit clothed humans obtained from normals.
\newblock In {\em CVPR}, 2022.

\bibitem{xu2020ghum}
Hongyi Xu, Eduard~Gabriel Bazavan, Andrei Zanfir, William~T Freeman, Rahul
  Sukthankar, and Cristian Sminchisescu.
\newblock Ghum \& ghuml: Generative 3d human shape and articulated pose models.
\newblock In {\em CVPR}, 2020.

\bibitem{yi2022mime}
Hongwei Yi, Chun-Hao~P. Huang, Shashank Tripathi, Lea Hering, Justus Thies, and
  Michael~J. Black.
\newblock {MIME}: Human-aware {3D} scene generation.
\newblock In {\em CVPR}, 2023.

\bibitem{yi2022generating}
Hongwei Yi, Hualin Liang, Yifei Liu, Qiong Cao, Yandong Wen, Timo Bolkart,
  Dacheng Tao, and Michael~J Black.
\newblock Generating holistic 3d human motion from speech.
\newblock In {\em CVPR}, 2023.

\bibitem{zanfir2020weakly}
Andrei Zanfir, Eduard~Gabriel Bazavan, Hongyi Xu, William~T Freeman, Rahul
  Sukthankar, and Cristian Sminchisescu.
\newblock Weakly supervised 3d human pose and shape reconstruction with
  normalizing flows.
\newblock In {\em ECCV}, 2020.

\bibitem{zeiler2014visualizing}
Matthew~D Zeiler and Rob Fergus.
\newblock Visualizing and understanding convolutional networks.
\newblock In {\em ECCV}, 2014.

\bibitem{zhang2021pymaf}
Hongwen Zhang, Yating Tian, Xinchi Zhou, Wanli Ouyang, Yebin Liu, Limin Wang,
  and Zhenan Sun.
\newblock Pymaf: 3d human pose and shape regression with pyramidal mesh
  alignment feedback loop.
\newblock In {\em ICCV}, 2021.

\bibitem{zhang2020object}
Tianshu Zhang, Buzhen Huang, and Yangang Wang.
\newblock Object-occluded human shape and pose estimation from a single color
  image.
\newblock In {\em CVPR}, 2020.

\bibitem{zhou2019continuity}
Yi Zhou, Connelly Barnes, Jingwan Lu, Jimei Yang, and Hao Li.
\newblock On the continuity of rotation representations in neural networks.
\newblock In {\em CVPR}, 2019.

\end{thebibliography}
}

\end{document}